\documentclass{article}

\usepackage{nips15submit_e,times}

\usepackage{graphicx}
\usepackage{amsmath,amssymb,amsfonts,amsbsy,amsthm}
\usepackage{latexsym}

\usepackage{algorithm,algpseudocode}
\usepackage{caption}
\usepackage{subcaption}

\input{def.set}

\newcommand{\Prob}{\mathbb P}
\newcommand{\Ind}{\mathbb I}

\newcommand{\algrule}[1][.2pt]{\par\vskip.5\baselineskip\hrule height #1\par\vskip.5\baselineskip}

\nipsfinalcopy 

\title{Gaussian Copula Variational Autoencoders for Mixed Data}

\author{Suwon Suh and Seungjin Choi \\
Department of Computer Science and Engineering \\
Pohang University of Science and Technology \\
77 Cheongam-ro, Nam-gu, Pohang 37673, Korea \\
\texttt{\{caster,seungjin\}@postech.ac.kr}
}

\begin{document}

\maketitle

\begin{abstract}
The variational autoencoder (VAE) is a generative model with continuous latent variables where a pair of
probabilistic encoder (bottom-up)  and decoder (top-down) is jointly learned by stochastic gradient variational Bayes.
We first elaborate Gaussian VAE, approximating the local covariance matrix of the decoder as an outer product of the principal direction
at a position determined by a sample drawn from Gaussian distribution.
We show that this model, referred to as VAE-ROC, better captures the data manifold, compared to the standard Gaussian VAE
where independent multivariate Gaussian was used to model the decoder.
Then we extend the VAE-ROC to handle mixed categorical and continuous data.
To this end, we employ Gaussian copula to model the local dependency in mixed categorical and continuous data,
leading to {\em Gaussian copula variational autoencoder} (GCVAE).
As in VAE-ROC, we use the rank-one approximation for the covariance in the Gaussian copula, to capture the local
dependency structure in the mixed data.
Experiments on various datasets demonstrate the useful behaviour of VAE-ROC and GCVAE, compared to the standard VAE.
\end{abstract}

\section{Introduction}
\label{sec:introduction}

The variational autoencoder (VAE) \cite{KingmaDP2014iclr} is a deep probabilistic model with great popularity,
where a top-down probabilistic decoder is paired with a bottom-up probabilistic encoder
that is designed to approximate the posterior inference over continuous latent variables.
When the VAE is applied to a certain task in practice, the first thing to be determined is to assign a right probability distribution to each
dimension of the input data.
For example, we assign multivariate Gaussian distribution to real-valued attributes or independent Bernoulli distribution 
to each binary attribute of the input data.
This is a critical but cumbersome step to successfully train a VAE on a dataset in practical applications.

The Gaussian VAE makes assumption that the probabilistic decoder is modelled as multivariate Gaussian distribution with diagonal covariance matrix.
In such a case, it was observed that the Gaussian VAE was not successful in capturing a fruitful representation of real-valued MNIST dataset.
This enforces us to binarize the data, followed by training the Bernoulli VAE on.
In other words, the independent multivariate Gaussian assumption on the decoder in the Gaussian VAE limits its capability to capture
the data manifold.
To this end, we extends the Gaussian VAE, approximating the local covariance matrix of the decoder as an outer product of the principal direction
at a position determined by a sample drawn from Gaussian distribution.
A closely-related work is the nonlocal estimation of manifold structure \cite{BengioY2006neco} where local manifold structure is determined by
modelling linear tangent space as a nonlinear function (for instance, MLP) of data, which can model smoothly changing tangent spaces. 
While this work chooses the membership of tangent space with $k$-NN, our model chooses its membership by setting local dependency 
as a function of latent variables, which is efficiently inferred by the recognition network.
We show that this model, referred to as VAE-ROC, better captures the data manifold, compared to the standard Gaussian VAE
where independent multivariate Gaussian was used to model the decoder.

\if(0)
It is observed that the Gaussian VAE with diagonal covariance matrix is fundamentally flaw to capture data manifold due to diagonal covariance matrix. 
Manifold in high dimensional input space is a set of regions with high density of data, which can be highly complex as a whole but 
\textit{locally linear in its tangent space}. With this clue, we can improve the performance of Gaussian VAE with non diagonal covariance matrix 
that reflect locally linear tangent space as outer product of principal direction.
In other words, rather than simply generating reconstruction in VAE, we additionally generate locally linear tangent space that the reconstruction resides. 
This can be interpreted as introducing Maharanobis distance with non-diagonal covariance matrix in Neural Network community 
and can be also interpreted as infinite mixture of principal component analyzer in probabilistic graphical model community.

Mainly, manifold learning have been tackled with graph based models that tried to find a manifold that preserves a certain relationship between instances in close distance, which depends on different criteria per model \cite{Cox,TenenbaumJB2000science,RoweisST2000science,MaatenL2008jmlr}. On the other hand, there have been a series of effort to solve manifold learning with mixture of Gaussians, where the main problem is to find a global coordinate that is most agreeable by local coordinates of Gaussians that overlaps each other \cite{BrandM2002nips, RoweisST2001nips}. In our model, VAE-ROC, it a point in global coordinate in input space has membersihps of infinite number of mixtures indexed by a value in latent space, which is slowly changing its local cooridnate along with changing its membership along with a smooth change on latent space.  Moreover, this work is closely related to Bengio's work where he tried to find a local manifold structure by modelling linear tangent space as a nonlinear function (MLP) of data, which can model smoothly changing tangent spaces \cite{BengioY2006neco}. While this work chooses the membership of tangent space with K Nearest Neighborhood (KNN), our work chooses its membership by setting local dependency as a function of latent variables, which is efficiently inferred by recognition network.
\fi

\if(0) %
Tangent propagation is also closely related model in supervised setting, where it tried to enhance supervised network by making it invariant to the local transformations, which constitute basis of the local tangent space, are manually constructed \cite{PatriceS1992nips}.
\fi

We extends this idea to model tangent space on mixed variate data using copula. Copula is a way of modeling multivariate distribution from bottom-up way. Specifically, we can model it with two steps: (1) modelling marginal distribution for each input dimension. (2) modeling dependency structure among them with copula \cite{NelsenRB2006book}.  Because each random variable can be transformed to the uniform random variable on [0, 1] with the corresponding cumulative distribution function (CDF), we can handle dependency among mixed variate data easily using copulas.
There are three points we must mention before explaining how we address modeling tangent space in mixed variate data:  (1) modeling mixed variate using copula, (2) parsimonious parameterization using copula, (3) mixture models in copula.

 First, in earlier days, Everitt introduced a latent Guassian variable for each categorical variable and model them with observed continuous variables in form of multivariate Normal distribution \cite{EverittBS88spl}. Hoff proposed extended rank likelihood  for mixed variate data, where it ignores the form of margins by calculating likelihood of a set of projected rank preserved data on unit hypercube \cite{HoffPD2007aas}. Continuous Extension (CE)  is a way of incorporating discrete variable into copula by introducing jittered uniform variable to a discrete variable \cite{DenuitM2005jma}. In GCVAE, we follows the idea of simulated likelihood by margin out jittered uniform variable with one sample \cite{MadsenL2011biometrics}. GCVAE cannot use extended rank likelihood because it models copula as well as marginal distributions.

 Second, for parsimonious parameterization of dependency in copula, Copula Gaussian Graphical Models (CGGM) introduce sparsity in precision matrix to reflect conditional independence across input dimensions \cite{LiuH2009jmlr}. Murray proposed a latent factor model for normal scores, which assume structure in correlation matrix in Gaussian copula \cite{MurrayJS2013jasa}.  GCVAE shares the idea of Bayesian Gaussian Copula Factor Model but introduce simple enough correlation across input dimension by normalizing rank one covariance matrix that consists of outer product of principal direction vector with isotropic noise.

 Finally, Gaussian copula can be extended to a finite mixture model that takes a subset of data that shows similar dependency among dimensions into one mixture component \cite{MarbacM2015arxiv}.  GCVAE extends this idea into infinite mixture model each component of which is indexed by a vector on latent space. Along with smooth change of latent space, the membership of each mixture component is smoothly changing (marginal distribution is smoothly changing) as well as local dependency across input dimension in the mixture component (parsimonious parameter of copula is smoothly changing).

In this paper we address two issues on the VAE: (1) better representation learning by capturing local principal directions at different positions
which vary along the data manifold; (2) managing mixed continuous and discrete data in practice.

\section{Background}
\label{sec:background}

In this section, we briefly review the variational autoencoder \cite{KingmaDP2014iclr} and copulas \cite{NelsenRB2006book}
on which we base our development of Gaussian copula variational autoencoder.

\subsection{Variational Autoencoders}
\label{subsec:vae}

The VAE is a probabilistic model for a generative process of observed variables $\bx \in \Real^{D}$
with continuous latent variables $\bz \in \Real^{K}$, which
pairs a top-down probabilistic decoder $p_{\theta}(\bx | \bz)$ with a bottom-up probabilistic encoder $q_{\phi}(\bz | \bx)$
that approximates the posterior inference over latent variables $\bz$.
In the case where both decoder and encoder are multivariate Gaussians with diagonal covariance matrices,
the decoder of Gaussian VAE is described by
\be
\label{eq:vae1}
p_{\theta}(\bz) & = & \calN (\bz \,|\, 0, \bI), \nonumber \\
p_{\theta}(\bx | \bz) & = & \calN (\bx \,|\, \bmu, \bLambda_{\sigma}),
\ee
where mean vector $\bmu$ and diagonal covariance matrix $\bLambda_{\sigma}=\mbox{diag} \big(\bsigma^2 \big)$
with $\bsigma^2=[\sigma_1^2,\ldots,\sigma_D^2]^{\top}$
are computed by MLPs:
\be
\label{eq:vae2}
\bmu & = & \bW_{\mu} \bh + \bb_{\mu}, \nonumber \\
\log \bsigma^2 & = & \bW_{\sigma} \bh + \bb_{\sigma}, \nonumber \\
\bh & = & \tanh(\bW_h \bz + \bb_h).
\ee
This is an example of MLP with a single hidden layer. Certainly, multiple hidden layers can be considered.
In order to emphasize the dependency of $\bmu$ and $\bsigma^2$ to the latent variable $\bz^{(n)}$,
we also use $\bmu(\bz^{(n)}) = \mbox{MLP}_{\mu} (\bz^{(n)})$ and
$\bsigma^2(\bz^{(n)})  = \mbox{MLP}_{\sigma} (\bz^{(n)})$.
The set of parameters, $\theta$, for the decoder, is given by
\bee
\theta=\{\bW_{\mu}, \bW_{\sigma},\bW_{h},\bb_{\mu}, \bb_{\sigma}, \bb_{h} \}.
\eee
The encoder of Gaussian VAE is described by
\be
\label{eq:vae3}
q_{\phi}(\bz | \bx) & = & \calN(\bz \,|\, \boldeta, \bLambda_{\tau}),
\ee
where $\bLambda_{\tau} = \mbox{diag} \big( \btau^2 \big)$ and parameters are also computed by MLPs:
\be
\label{eq:vae4}
\boldeta & = & \bV_{\eta} \by + \bb_{\eta}, \nonumber \\
\log \btau^2 & = & \bV_{\tau} \by + \bb_{\tau}, \nonumber \\
\by & = & \tanh( \bV_{y} \bx + \bb_{y} ).
\ee
The set of parameters, $\phi$, for the encoder, is given by
\bee
\phi = \{ \bV_{\eta}, \bV_{\tau}, \bV_{y}, \bb_{\eta}, \bb_{\tau}, \bb_{y} \}.
\eee

Suppose that we are given a training dataset $\bX=[\bx^{(1)},\ldots,\bx^{(n)}]$ consisting of $N$ i.i.d. observations,
each of which is a $D$-dimensional random vector.
The marginal log-likelihood of $\bX$ is a sum over the marginal log-likelihood of individual observations:
\bee
\log p_{\theta} (\bX) = \sum_{n=1}^{N} \log p_{\theta}(\bx^{(n)}),
\eee
where the single factor of the marginal log-likelihood, $\log p_{\theta}(\bx^{(n)})$, is described by its variational lower-bound:
\bee
\lefteqn{ \log p_{\theta}(\bx^{(n)}) } \\
& = &  \log \int p_{\theta}(\bx^{(n)}, \bz^{(n)}) d\bz^{(n)} \\
& \geq & q_{\phi}(\bz^{(n)}|\bx^{(n)}) \log \left[ \frac{ p_{\theta}(\bx^{(n)} | \bz^{(n)}) p_{\theta}(\bz^{(n)})}{q_{\phi}(\bz^{(n)}|\bx^{(n)})} \right] d\bz^{(n)} \\
& = & \calF(\theta,\phi; \bx^{(n)}).
\eee
The variational lower-bound, $\calF(\theta,\phi; \bx^{(n)})$, on the marginal log-likelihood of observation $\bx^{(n)}$ is given by
\be
\label{eq:elbo_vae}
\calF(\theta,\phi; \bx^{(n)})  & = &  \E_{q_{\phi}(\bz^{(n)}|\bx^{(n)})} \left[ \log p_{\theta}(\bx^{(n)} | \bz^{(n)}) \right] \nonumber \\
& - &\mbox{KL}\left[ q_{\phi}(\bz^{(n)}|\bx^{(n)}) \,\|\, p_{\theta}(\bz^{(n)}) \right],
\ee
where $\mbox{KL}[ q \,\|\, p ]$ is the KL-divergence between the distributions $q$ and $p$.
The second term in (\ref{eq:elbo_vae}) is analytically computed and the first term is calculated by
the stochastic gradient variational Bayes (SGVB) where the Monte Carlo estimates are performed with the reparameterization trick:
\bee
\lefteqn{ \E_{q_{\phi}(\bz^{(n)}|\bx^{(n)})} \left[ \log p_{\theta}(\bx^{(n)} | \bz^{(n)}) \right] } \\
& \approx & \frac{1}{L} \sum_{l=1}^{L} \log p_{\theta}(\bx^{(n)} | \bz^{(n,l)}),
\eee
where $\bz^{(n,l)} = \boldeta (\bx^{(n)}) + \btau(\bx^{(n)}) \odot \bepsilon^{(l)}$ ($\odot$ represents
the elementwise product) and $\bepsilon^{(l)} \sim \calN (0,\bI)$.
A single sample is often sufficient to form this Monte Carlo estimates in practice, thus, in this paper we simply use
\bee
\lefteqn{ \E_{q_{\phi}(\bz^{(n)}|\bx^{(n)})} \left[ \log p_{\theta}(\bx^{(n)} | \bz^{(n)}) \right] } \\
& \approx & \log p_{\theta}(\bx^{(n)} | \bz^{(n)}),
\eee
where $\bz^{(n)} = \boldeta (\bx^{(n)}) + \btau(\bx^{(n)}) \odot \bepsilon$ denotes a single sample with abuse of notation.

\subsection{Copulas}
\label{subsec:copula}

A $D$-dimensional {\em copula} $C$ is a distribution function on unit cube $[0, 1]^D$ with each univariate marginal
distribution being uniform on $[0,1]$. A classical result of Sklar \cite{SklarA59} relates a joint cumulative distribution function (CDF),
$F(x_1,\ldots,x_D)$, of $D$ random variables $\{x_i\}_{i=1}^{D}$, to a copula function $C(F_1(x_1),\ldots,F_D(x_D))$ via
univariate marginal CDFs, $F_i$, i.e.,
\be
\label{eq:sklar}
F(x_1,\ldots,x_D) = C \big(F_1(x_1),\ldots,F_D(x_D) \big).
\ee
If all $F_i$ are continuous, then the copula satisfying (\ref{eq:sklar}) is unique and is given by
\be
C(u_1,\ldots,u_D) = F \big( F_1^{-1}(u_1), \ldots, F_D^{-1}(u_D) \big),
\ee
for $u_i = F_i(x_i) \in [0,1], i=1,\ldots,D$ where $F_i^{-1}(u_i) = \inf\{ x_i \,|\, F_i(x_i) \geq u_i\}, i=1,\ldots,D.$
Otherwise it is uniquely determined on $\calR(F_1) \times \cdots \times \calR(F_D)$ where $\calR(F_i)$ is
the range of $F_i$.
The relation in (\ref{eq:sklar}) can also be represented in terms of the joint
probability density function (PDF), $p(x_1,\ldots,x_D) = p(\bx), $ when $x_i$'s are continous:
\be
\label{eq:sklar1}
p(x_1,\ldots,x_D) = c \big(F_1(x_1),\ldots,F_D(x_D) \big) \prod_{i=1}^{D} p_i(x_i),
\ee
where $c(u_1,\ldots,u_D) = \frac{\partial C(u_1,\ldots,u_D)}{\partial u_1 \cdots \partial u_D}$ is the
{\em copula density} and $p_i(x_i)$ are marginal PDFs.

The Gaussian copula with covariance matrix $\bSigma \in \Real^{D \times D}$, that we consider in this paper, is given by
\bee
\label{eq:gcopula}
C_{\Phi}(u_1,\ldots,u_D) = \Phi_{\Sigma} \big( \Phi^{-1}(u_1),\ldots,\Phi^{-1}(u_D) \,|\, \bSigma \big),
\eee
where $\Phi_{\Sigma}(\cdot \,|\, \bSigma)$ is the $D$-dimensional Gaussian CDF with covariance matrix $\bSigma$ with diagonal entries
being equal to one
and $\Phi(\cdot)$ is the univariate standard Gaussian CDF.
The Gaussian copula density is given by
\bee
\label{eq:gcd}
c_{\Phi}(u_1,\ldots,u_D)  & = & \frac{\partial^{D} C_{\Phi} (u_1,\ldots,u_D) }{\partial u_1 \cdots \partial u_D} \nonumber \\
& = & |\bSigma|^{-\frac{1}{2}} \exp \left\{ -\frac{1}{2} \bq^{\top} (\bSigma^{-1} - \bI) \bq \right\},
\eee
where $\bq=[q_1,\dots,q_D]^{\top}$ with {\em normal scores} $q_i = \Phi^{-1}(u_i)$ for $i=1,\ldots,D$.
Invoking (\ref{eq:sklar1}) with this Gaussian copula density, the joint density function is written as:
\be
\label{eq:gcd1}
p(\bx) =  |\bSigma|^{-\frac{1}{2}} \exp \left\{ -\frac{1}{2} \bq^{\top} (\bSigma^{-1} - \bI) \bq \right\}
\prod_{i=1}^{D} p_i(x_i).
\ee

If $x_i$'s are discrete, the copula $C_{\Phi}(u_1,\ldots,u_D)$ are uniquely determined on the range of $F_1 \times \cdots \times F_D$.
In such a case, the joint probability mass function (PMF) of $x_1,\ldots, x_D$ is given by
\be
\label{eq:pmf_gcd}
p(x_1,\ldots, x_D) & = & \sum_{j_1=1}^{2} \cdots \sum_{j_D=1}^2 (-1)^{j_1+\cdots j_D} \nonumber  \\
& & \hspace*{-.3in} \Phi_{\Sigma} \big( \Phi^{-1}(u_{1,j_1}),\ldots, \Phi^{-1}(u_{D,j_D}) \big),
\ee
where $u_{i,1}=F_i(x_i^{-})$, the limit of $F_i(\cdot)$ at $x_i$ from the left, and $u_{i,2}=F_i(x_i)$.
The PMF requires the evaluation of $2^D$ terms, which is not manageable even for a moderate value of $D$ (for instance, $D \geq 5$).
A continuous extension (CE) of discrete random variables $x_i$ \cite{DenuitM2005jma,MadsenL2011biometrics}
avoids the $D$-fold summation in (\ref{eq:pmf_gcd}), associating a continuous random variable $x_i^{\ast} = x_i - v_i$
with the integer-valued $x_i$, where $v_i$ is uniform on [0,1] and is independent of $x_i$ as well as of $\rho_j$ for $j \neq i$.
Continuous random variables $x_i^{\ast}$ produced by jittering $x_i$ yields the CDF and PDF given by
\bee
\label{eq:ce_cdf}
F_i^{\ast}(\xi) & = & F_i([\xi]) + (\xi-[\xi]) \Prob \left( x_i = [\xi+1] \right), \\
\label{eq:ce_pdf}
p_i^{\ast}(\xi) & = & \Prob \left( x_i = [\xi+1] \right),
\eee
where $[\xi]$ represents the nearest integer less than or equal to $\xi$.
The joint PDF for jittered variables $x_i^{\ast}$ is determined by substituting $F_i^{\ast}$ and $p_i^{\ast}$
into (\ref{eq:gcd1}). Then, averaging this joint PDF over the jitters $\bv=[v_1,\ldots,v_D]^{\top}$
lead to the joint PFM for $x_1,\ldots,x_D$:
{\small
\bee
\lefteqn{ p(x_1,\ldots,x_D) } \\
& = & \hspace*{-.1in} \E_{\bv} \left[
|\bSigma|^{-\frac{1}{2}} \exp \left\{ -\frac{1}{2} \bq^{\ast \top} (\bSigma^{-1} - \bI) \bq^{\ast} \right\}
 \prod_{i=1}^{D} p_i^{\ast}(x_i - v_i)  \right],
\eee}
where
\bee
\bq^{\ast} = \big[\Phi^{-1}(F_1^{\ast}(x_1 - v_1)), \ldots, \Phi^{-1}(F_D^{\ast}(x_D - v_D)) \big]^{\top}.
\eee

Given a set of $N$ data points $\bx^{(n)}$, in the case of Gaussian copula, the log-likelihood is given by invoking (\ref{eq:sklar1})
and (\ref{eq:gcd1})
\bee
\lefteqn{ \sum_{n=1}^{N} \log p(x^{(n)}_{1}, \ldots, x^{(n)}_{D}) } \\
&=&  -\frac{N}{2} \log |\bSigma| + \frac{1}{2} \sum_{n=1}^{N}  \bq^{(n)\top} (\bI-\bSigma^{-1}) \bq^{(n)} \\
& & + \sum_{n=1}^{N} \sum_{i=1}^{D} \log p_i(x^{(n)}_{i}).
\eee
Denote by $\vartheta_i$ the parameters that involve specifying the marginal PDF $p_i(\cdot)$.
Then the parameters $\{\vartheta_i\}$ and $\bSigma$ appearing in the Gaussian copula density are estimated by
the two-step method, known as {\em inference for margin} \cite{ChorosB2010lns}:
\bee
\widehat{\vartheta_i} & = & \argmax \sum_{n=1}^{N} \sum_{i=1}^{D} \log p_i(x^{(n)}_{i}; \vartheta_i), \\
\widehat{\bSigma} & = & \argmax \sum_{n=1}^{N} \log c_{\Phi} \big( u^{(n)}_{1}, \ldots, u^{(n)}_{D}; \widehat{\vartheta_i},\bSigma \big).
\eee

\section{VAE for Manifold Tangent Learning}
\label{sec:VAE-ROC}

We describe our model VAE-ROC in the perspective of manifold learning, making a simple modification in the Gaussian
VAE described in Section \ref{subsec:vae}.
Note that VAE-ROC is limited to continuous data only and its extension to mixed-data is presented in Section \ref{sec:gcvae}

As in VAE, VAE-ROC also constitutes a pair of probabilistic decoder and encoder.
The probabilistic encoder in VAE-ROC is the same as (\ref{eq:vae4}) but the decoder is slightly different from the VAE,
which is described below in detail.
In order to find the local principal direction at a specific location $\bmu(\bz)$, we use the following model for the probabilistic decoder
\be
\label{eq:vae_roc1}
p_{\theta}(\bx | \bz)
&=&  \mathcal{N} \left( \bmu, \omega \bI + \ba \ba^{\top} \right),
\ee
where the local covariance matrix is of the form $\omega \bI + \ba \ba^{\top} $
and each of $\bmu, \omega, \ba$ is parameterized by an individual MLP which takes $\bz$ as input.
For instance,
\be
\label{eq:vae_roc2}
p_{\theta}(\bz)  &=&  \mathcal{N}(0, \bI), \nonumber \\
\bmu &=& \bW_{\mu}\bh + \bb_{\mu}, \nonumber\\
\log \omega &=& \bw^{\top}_{\omega} \bh + b_{\omega}, \nonumber\\
\ba &=& \bW_{a}\bh + \bb_{a}, \nonumber\\
\bh &=& \tanh (\bW_{h}\bz + \bb_{h}).
\ee

In fact, VAE-ROC, can be viewed as an infinite mixture of probabilistic principal component analyzers.
Introducing a dummy Gaussian latent variable $s \sim \calN(0,1)$, the probability distribution over $\bx$ given $s$ and $\bz$ is given by
\bee
p(\bx | s, \bz) = \calN( \ba s + \bmu, \omega \bI).
\eee
The latent variable $\bz$ can be viewed as an indicator variable involving a local region to be approximated by a tangent plane..
However, it is not a discrete variable unlike the standard finite mixture model.
The variable $\bz$ is drawn from the standard multivariate Gaussian distribution and $\bmu(\bz) = \mbox{MLP}_{\mu} (\bz)$
determines a location where we approximate the local manifold structure as local principal direction $\ba$.
Besides this subtle difference, it can be viewed as the mixture of probabilistic principal component analyzers.
Marginalizing $s$ out yields the distribution for the decoder of VAE-ROC, given in (\ref{eq:vae_roc1}).

\begin{figure}[htp]
\begin{center}
\includegraphics[width=.7\columnwidth]{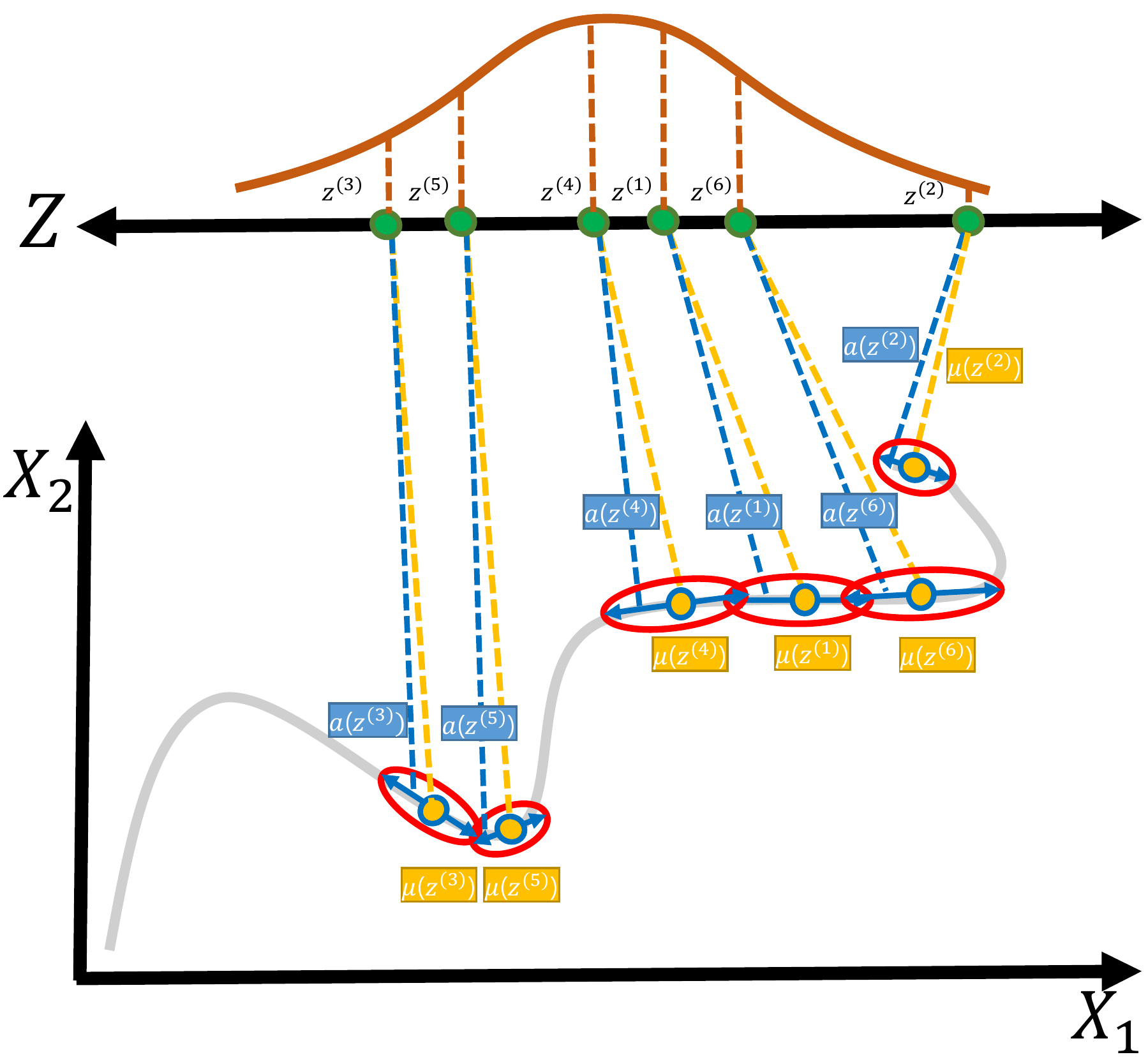}
\end{center}
\caption{Latent variables $\bz^{(n)}$ are sampled from the standard multivariate Gaussian distribution. 
These variables are fed into MLPs to determine the location $\bmu(\bz^{(n)})$ and the local principal direction $\ba(\bz^{(n)})$.
Data manifold is approximated by infinitely many local principal directions.}
\end{figure}

Given a training dataset $\bX=[\bx^{(1)},\ldots,\bx^{(n)}]$ consisting of $N$ i.i.d. observations,
each of which is a $D$-dimensional random vector, the variational lower-bound on the marginal log-likelihood of $\bX$ is given by
\be
\label{eq:elbo_vae_roc}
\calF(\theta,\phi; \bx^{(n)})  & = &  \E_{q_{\phi}(\bz^{(n)}|\bx^{(n)})} \left[ \log p_{\theta}(\bx^{(n)} | \bz^{(n)}) \right] \nonumber \\
& - &\mbox{KL}\left[ q_{\phi}(\bz^{(n)}|\bx^{(n)}) \,\|\, p_{\theta}(\bz^{(n)}) \right],
\ee
where the decoder $p_{\theta}(\bx^{(n)} | \bz^{(n)})$ is given in (\ref{eq:vae_roc1}) and the encoder
$q_{\phi}(\bz^{(n)}|\bx^{(n)})$ is described in (\ref{eq:vae3}) and (\ref{eq:vae4}).
The second term in (\ref{eq:elbo_vae_roc}) can be analytically calculated as:
\bee
\lefteqn{ \mbox{KL} \left[ q_{\phi}(\bz^{(n)} | \bx^{(n)}) \,\|\,p_{\theta}(\bz^{(n)}) \right] } \\
& = & -\frac{1}{2} \sum_{k=1}^{K} \left( 1 + 2\log \tau_{k}^{(n)} - \left( \tau_{k}^{(n)} \right)^2 - \left( \eta_{k}^{(n)} \right)^2 \right),
\eee
where the superscript $^{(n)}$ is used for $\eta_k$ and $\tau_k$ to reflect their dependence on $\bx^{(n)}$.
The first term in (\ref{eq:elbo_vae_roc}) is calculated by
the stochastic gradient variational Bayes (SGVB) where the Monte Carlo estimates are performed with the reparameterization trick \cite{KingmaDP2014iclr}:
\bee
\lefteqn{ \E_{q_{\phi}(\bz^{(n)}|\bx^{(n)})} \left[ \log p_{\theta}(\bx^{(n)} | \bz^{(n)}) \right] } \\
& \approx & \frac{1}{L} \sum_{l=1}^{L} \log p_{\theta}(\bx^{(n)} | \bz^{(n,l)}),
\eee
where $\bz^{(n,l)} = \boldeta^{(n)} + \btau^{(n)} \odot \bepsilon^{(l)}$ ($\odot$ represents
the elementwise product) and $\bepsilon^{(l)} \sim \calN (0,\bI)$.
A single sample is often sufficient to form this Monte Carlo estimates in practice, thus, in this paper we simply use
\bee
\E_{q_{\phi}(\bz^{(n)}|\bx^{(n)})} \left[ \log p_{\theta}(\bx^{(n)} | \bz^{(n)}) \right]
 \approx  \log p_{\theta}(\bx^{(n)} | \bz^{(n)}),
\eee
where $\bz^{(n)} = \boldeta^{(n)} + \btau^{(n)} \odot \bepsilon$.

Denote by $\bX_M=\left\{\bx^{(m)} \right\}_{m=1}^{M}$ is a minibatch of size $M$, which consists of $M$ randomly drawn samples
from the full dataset $\bX$ with $N$ points.
Then an estimator of the variational lower-bound is constructed by
\be
\label{eq:elbo_vae_roc1}
\calF(\theta,\phi; \bX) \approx \frac{N}{M} \sum_{m=1}^{M} \calF(\theta,\phi; \bx^{(m)}),
\ee
where
{\small
\bee
\lefteqn{  \calF(\theta,\phi; \bx^{(m)}) } \\
&& = -\frac{D}{2} \log 2\pi  - \frac{1}{2} \log \left|\omega^{(m)}\bI + \ba^{(m)}{\ba^{(m)}}^{\top} \right|\\
&& -\frac{1}{2} \widetilde{\bx}^{(m) \top} \left(\omega^{(m)}\bI + \ba^{(m)}{\ba^{(m)}}^{\top} \right)^{-1} \widetilde{\bx}^{(m)} \\
&& +\frac{1}{2}\sum_{k=1}^{K} \left(1 + 2\log {\tau_{k}^{(m)}} - \left( \tau_{k}^{(m)} \right)^2 - \left( \eta_{k}^{(m)} \right)^2 \right),
\eee
where $\widetilde{\bx}^{(m)} = \bx^{(m)}-\bmu^{(m)}$.
Applying the Sherman-Morrison formula to $ \left(\omega^{(m)}\bI + \ba^{(m)}{\ba^{(m)}}^{\top} \right)^{-1}$ leads to
\be
\label{eq:elbo_vae_roc2}
\lefteqn{  \calF(\theta,\phi; \bx^{(m)}) } \nonumber \\
&&= -\frac{D}{2} \log 2\pi - \frac{D}{2} \log \omega^{(m)} - \frac{1}{2} \log \left( 1 + \frac{ {\ba^{(m)}}^{\top} \ba^{(m)} }{\omega} \right) \nonumber \\
&&= -\frac{1}{2} \widetilde{\bx}^{(m) \top} \left( \frac{1}{\omega^{(m)}} \bI - \frac{  \ba^{(m)} {\ba^{(m)}}^{\top} }
{ {\omega^{(m)}}^2 + \omega^{(m)}{\ba^{(m)}}^{\top} \ba^{(m)} } \right) \widetilde{\bx}^{(m)} \nonumber \\
&& +\frac{1}{2}\sum_{k=1}^{K} \left(1 + 2\log {\tau_{k}^{(m)}} - \left( \tau_{k}^{(m)} \right)^2 - \left( \eta_{k}^{(m)} \right)^2 \right).
\ee}

With the maximization of the variational lower-bound, we consider two regularization terms, each of which is explained below:
\begin{itemize}
\item
\textbf{Locality Regularization:} The first regularization is the $\ell_p$ regularization applied to the local principal direction $\ba^{(m)}$ to enforce a bound on its length.
It was observed in our experiments that both $\ell_1$ and $\ell_2$ norm worked well where $\ell_1$ norm gave more spare local principal direction.
\item
\textbf{Rank Regularization:} Denote by  $\widetilde{p}(\bx)$ the data distribution.
The aggregated posterior $q_{\phi}(\bz) = \int q_{\phi}(\bz| \bx) \widetilde{p}(\bx) d\bx$ on latent variables $\bz$ is expected to match
the Gaussian prior distribution $p_{\theta}(\bz) = \calN(0,\bI)$.
Adversarial autoencoder \cite{MakhzaniA2016iclr} is a recent example which incorporates this regularization.
In this paper, we take a different approach, which is close to the idea used in learning neural networks to estimate the distribution function \cite{MagdonIsmailM98nips}.
Note that the random variable $\Phi_{I}(\bz)$ (where $\Phi_{i}(\cdot)$ denotes the standard Gaussian CDF) is uniform in [0,1].
Thus, we consider the following penalty function:
\bee
\sum_{m=1}^{M}  \frac{1}{2}\left(u^{(m)} - \Phi_{I}(\bz^{(m)}) \right)^2,
\eee
where $u^{(m)}$ are randomly drawn samples from uniform distribution in [0,1] and
$\bz^{(m)} \sim q_{\phi}(\bz)$. Samples $u^{(m)}$ and $\bz^{(m)}$ are sorted in ascending order to relate
$ \Phi_{I}(\bz^{(m)})$ to $u^{(m)}$.
\end{itemize}
Thus our regularized variational lower-bound is given by
\be
\label{eq:elbo_vae_roc_reg}
\widetilde{\calF}(\theta,\phi; \bX) & = & \frac{N}{M} \sum_{m=1}^{M} \left( \calF(\theta,\phi; \bx^{(m)})
- \lambda_a \|\ba^{(m)}\|_{p}^{2} \right) \nonumber \\
&- &  \frac{N}{M} \frac{\lambda_r}{2} \sum_{m=1}^{M}  \left(u^{(m)} - \Phi_{I}(\bz^{(m)}) \right)^2.
\ee
With the random minibatch $\bX_M$ of $M$ datapoints, this regularized variational lower-bound in (\ref{eq:elbo_vae_roc_reg}) is maximized
to determine parameters $\theta$ and $\phi$. This training is repeated with different minibatches until the convergence of parameters is achieved.

\section{Gaussian Copula VAE}
\label{sec:gcvae}

\label{subsec:GCoupulaVAE}
\begin{figure}[th]
    \centering
        \includegraphics[width=0.50\textwidth]{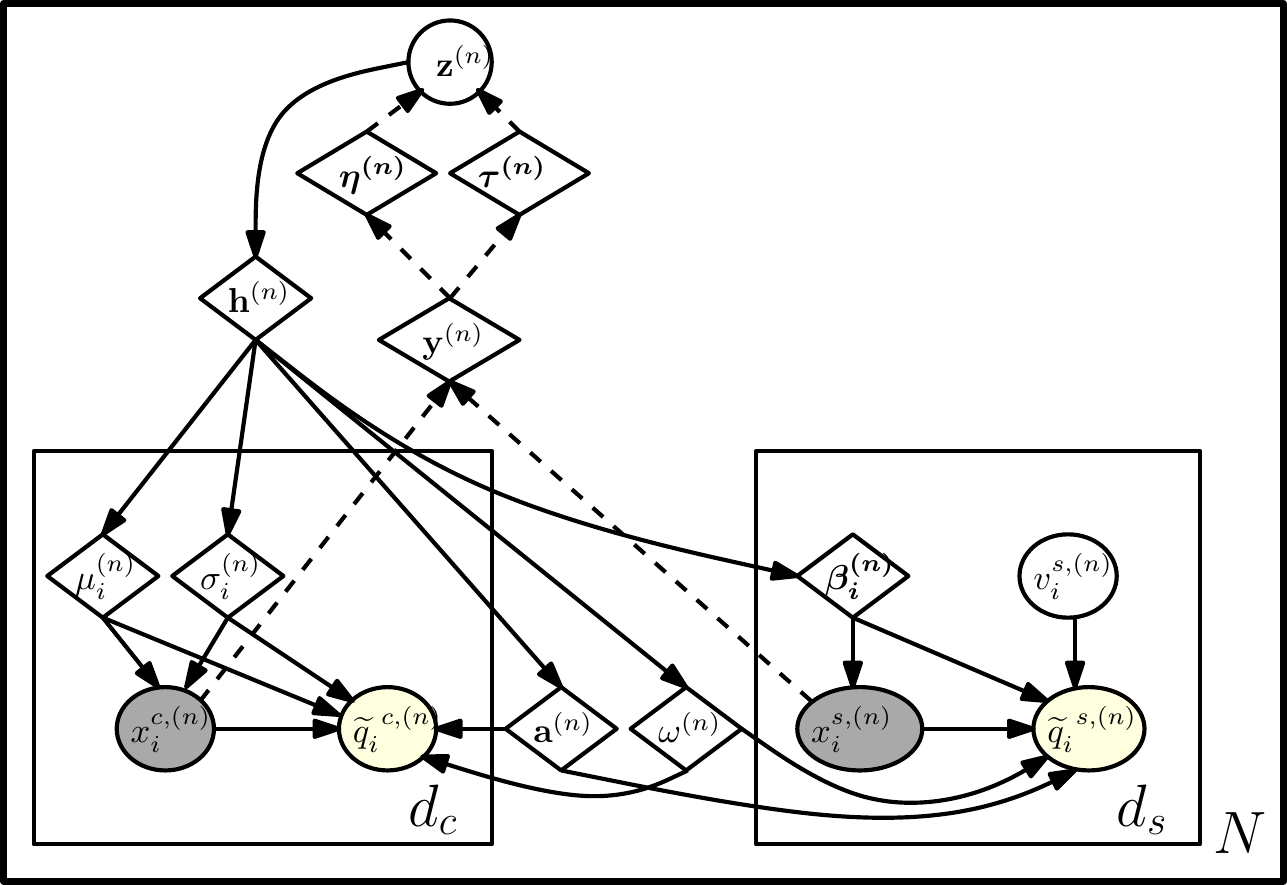}
        \caption{Gaussian Copula VAE with continuous random variables and discrete random variables: Circles denotes variables and diamonds denotes units consisting neural networks.}
        \label{fig:GaussianCopulaVAE}
\end{figure}

In this section we present a {\em Gaussian copula variational autoencoder} (GCVAE)
where we extend the VAE-ROC (described in the previous section)
developed for continuous data to handle mixed continuous and discrete data, employing Gaussian copula
explained in Section \ref{subsec:copula}.

Suppose we are given a set of $N$ data points, each of which is $D$-dimensional vector, i.e, $\bX = \{\bx^{(1)}, \ldots, \bx^{(N)}\}$.
We use superscripts $c$ and $d$ to indicate variables associated with continuous and discrete data, respectively,
i.e., $\bx^{(n)} =  [\bx^{c,(n)}; \bx^{s,(n)} ]$, where the semicolon is used to represent stacking vectors in a column.
Denote by $\bx^{c,(n)} \in \Real^{d_c}$ the vector which collects continuous attributes
and by $\bx^{s,(n)} \in \Real^{d_s}$ the vector containing discrete attributes.
The $i$th entry of $\bx^{s,(n)}$ is represented by $x^{s,(n)}_{i} \in \{1,\ldots, J\}$.

In principle, GCVAE allows for various continuous or discrete distributions together.
In this paper, however, we assume Gaussian distributions for continuous variables
and categorical distributions for discrete variables. That is,
\bee
p_i(x_i^{c}) & = & \calN( \mu_i, \sigma_i^2 ), \\
p_i(x_i^{s}) & = & \prod_{j=1}^{J} \beta_{i,j}^{\Ind ( x_i^{s}=j )},
\eee
where $\beta_{i,j }= \Prob (x_i^{s} = j )$ and $\Ind (\cdot)$ is the indicator function which yields 1 when the input argument is true
and otherwise 0.

In GCVAE, we use the Gaussian copula to model the probabilistic decoder $p_{\theta} (\bx | \bz)$, given by
\bee
p_{\theta} (\bx | \bz ) & = & p_{\theta}(\bx^{c}, \bx^{s} | \bz) \\
& = & \E_{\bv} \Big[  c_{\Psi}(\cdot) \prod_{i=1}^{d_c} p_i(x_i^c)
\prod_{i=1}^{d_s} p_i^{\ast} (x_i^s - v_i) \Big] \\
& \approx & c_{\Psi}(\cdot) \prod_{i=1}^{d_c} p_i(x_i^c)
\prod_{i=1}^{d_s} p_i^{\ast} (x_i^s - v_i) \\
& = & c_{\Psi}(\cdot) \prod_{i=1}^{d_c} \calN( \mu_i, \sigma_i^2)
\prod_{i=1}^{d_s} \prod_{j=1}^{J} \beta_{i,j}^{\Ind ( x_i^{s}=j )},
\eee
where the expectation is approximated using a single sample drawn from the uniform distribution on [0,1]
and the Gaussian copula density (see Appendix \ref{sec:gc} for proof) is given by
{\small
\bee
\lefteqn{ c_{\Psi} (\cdot)  } \\
& =  & c_{\Psi} \big( F_1(x_1^c),\ldots, F_{d_c}(x_{d_c}^c),  \\
& & F_1^{\ast}(x_1^s - v_1), \ldots, F_{d_s}^{\ast}(x_{d_s}^s - v_{d_s}) \big) \\
& = & \left( \prod_{i=1}^{D} \psi_i \right) | \bPsi |^{-\frac{1}{2}} \exp \left\{ -\frac{1}{2} \widetilde{\bq}^{\top}
\left( \bPsi^{-1} -  \bS^{-1} \right) \widetilde{\bq} \right\},
\eee}
where
{\small
\be
\label{eq:GCVAE_norm_score}
&&\widetilde{\bq}^{\top}  = \Big[\psi_1\Phi^{-1}(F_1(x_1^c)),\ldots, \psi_{d_c} \Phi^{-1}( F_{d_c}(x_{d_c}^c)), \\
&&\psi_{d_c+1}\Phi^{-1}( F_1^{\ast}(x_1^s - v_1)),\ldots, \psi_{d_c+d_s}\Phi^{-1}(F_{d_s}^{\ast}(x_{d_s}^s - v_{d_s})) \Big]~,\nonumber
\ee}
and the covariance matrix in Gaussian copula is of the form $\bPsi \triangleq \omega\bI + \ba\ba^{\top}$.
Diagonal entries of $\bPsi$ are denoted by $\psi_i^2$ and $\bS$ represents a diagonal matrix
with diagonal entries being $\psi_i^2$.
When $\bPsi = \bI$, the Gaussian copula density is equal to 1, i.e., GCVAE is identical to VAE.

As in the VAE-ROC, each of $\bmu, \bsigma^2, \ba, \omega, \bbeta$ is parameterized by an individual MLP which takes $\bz$ as input.
For instances,
\bee
p_{\theta}(\bz)  &=&  \mathcal{N}(0, \bI),\\
\bmu &=& \bW_{\mu}\bh + \bmu,\\
\log \bsigma^2 &=& \bW_{\sigma}\bh + \bb_{\sigma},\\
\beta_{i,j} &=& \frac{\exp \left\{\bw_{\beta,i,j}^{\top} \bh + b_{\beta,i,j} \right\}}
{\sum_{j^{\prime}=1}^{J}\exp \left\{\bw_{\beta,i,j^{\prime}}^{\top} \bh + b_{\beta,i,j^{\prime}} \right\}},\\
\log \omega &=& \bw^{\top}_{\omega}\bh + b_{\omega},\\
\ba &=& \bW_{a}\bh + \bb_{a},\\
\bh &=& \tanh(\bW_{h}\bz + \bb_{h}).
\eee
The probabilistic encoder $q_{\phi}(\bz | \bx)$ is parameterized, as in VAE or VAE-ROC, as described in (\ref{eq:vae3}) and (\ref{eq:vae4}).
As in VAE-ROC, GCVAE is learned by maximizing the reqularized variational lower-bound (\ref{eq:elbo_vae_roc_reg}),
where the lower-bound is given by
{\small
\be
\label{eq:elbo_gcvae}
\lefteqn{  \calF(\theta,\phi; \bx^{(m)}, \widetilde{\bq}^{(m)}) } \nonumber \\
&& =  \sum_{i=1}^{d_c+d_s} \log \psi_i  - \frac{1}{2} \log \left|\omega^{(m)}\bI + \ba^{(m)}{\ba^{(m)}}^{\top} \right| \nonumber  \\
&& -\frac{1}{2} \widetilde{\bq}^{(m) \top}\left[  \left(\omega^{(m)}\bI +
\ba^{(m)}{\ba^{(m)}}^{\top} \right)^{-1} - \bS^{-1} \right] \widetilde{\bq}^{(m)} \nonumber  \\
&& + \sum_{i=1}^{d_s}\sum_{j=1}^{J} \Ind ( x_i^{{s},(m)}=j ) \log \beta^{(m)}_{i,j}  \nonumber \\
&& - \sum_{i=1}^{d_c}\left( \frac{1}{2} \log (2\pi) +\log \sigma_i^{(m)}
+\frac{1}{2}\frac{ \left(x_i^{c,(m)}-\mu_i^{(m)} \right)^2}{ \left(\sigma_i^{(m)} \right)^2} \right)  \nonumber \\
&& +\frac{1}{2}\sum_{k=1}^{K} \left(1 + 2\log {\tau_{k}^{(m)}} - \left( \tau_{k}^{(m)} \right)^2 - \left( \eta_{k}^{(m)} \right)^2 \right).
\ee }
Note that normal scores  $\widetilde{\bq}$ are treated as constant, as in IFM, when the parameter estimation is carried out using
the variational lower-bound in (\ref{eq:elbo_gcvae}).
The last three terms  in (\ref{eq:elbo_gcvae}) are shared by the standard VAE, but the first two terms are introduced to model the local
dependency between marginals using Gaussian copula in GCVAE.
There are two remarks about this model.
\begin{itemize}
\item First of all, VAE-ROC can be treated as a special case of GCVAE with only continuous variables, each of which follows Gaussian distribution.
\item Second, just like VAE-ROC can be interpreted as infinite mixture of PPCA, GCVAE can be also interpreted as infinite mixture of Gaussian copula.
Unlike the mixture of Gaussian copula \cite{MarbacM2015arxiv}, the number of mixture component is infinite and the covariance matrix
of a Gaussian copula component is restricted by rank one structure with idiosyncratic noise.
\end{itemize}

\begin{algorithm}
  \caption{AEVB for GCVAE}\label{algo:gcvae}
  \begin{algorithmic}
    \Require
     $\bX = \left\{\bx^{(1)}, \cdots, \bx^{(N)} \right\}$.
    \algrule
    \Ensure $\theta$ and $\phi$
    \algrule
    \State Initialize $\theta, \phi$.
    \Repeat
    \State $\bX_{M} \leftarrow$ Random minibatch of $M$ data points drawn from $\bX$
    \State $\left\{ \bepsilon^{(m)} \right\}_{m=1}^{M} \leftarrow$ $M$ i.i.d. random samples drawn from $\calN(0, \bI)$.
    \State $\{ \ba^{(m)}, \omega^{(m)}, \{\mu_i^{(m)}, \sigma_i^{(m)}\}_{i=1}^{d_c}, \{\bbeta_i^{(m)}\}_{i=1}^{d_s}\}_{m=1}^{M} \leftarrow$  forward propagation from mini-batch (recognition network) via random source samples (generative network).
    \State $\{ \bv^{(m)} \}_{m=1}^{M} \leftarrow$ Sample jitter variables.
    \State $\{ \widetilde{\bq}^{(m)} \}_{m=1}^{M} \leftarrow$ update normal score with Eq. \ref{eq:GCVAE_norm_score}.
    \State $\bg \leftarrow \nabla_{\theta, \phi} \frac{N}{M}\sum_{m=1}^{M}{\calF(\btheta,\bphi;\bx^{(m)},\widetilde{\bq}^{(m)})}$ in Eq. \ref{eq:elbo_gcvae}.
    \State $(\theta, \phi) \leftarrow$ parameter updates using  ADAM \cite{KingmaDP2015iclr} with gradient $\bg$.
    \Until convergence of parameters $(\theta, \phi)$\\
  \end{algorithmic}
\end{algorithm}

\section{Experiments}
\label{sec:experiments}

In this section, we compare proposed models, VAE-ROC and GCVAE, with VAE on various continuous and mixed variate datasets respectively.
We implemented the models using Matlab and apply adaptive learning rate SGD algorithm, ADAM.

\subsection{Half circle dataset \& Half circle with wedges dataset}

\begin{figure}[t!]
\centering
    \begin{subfigure}[t]{0.4\textwidth}
        \includegraphics[width=\textwidth]{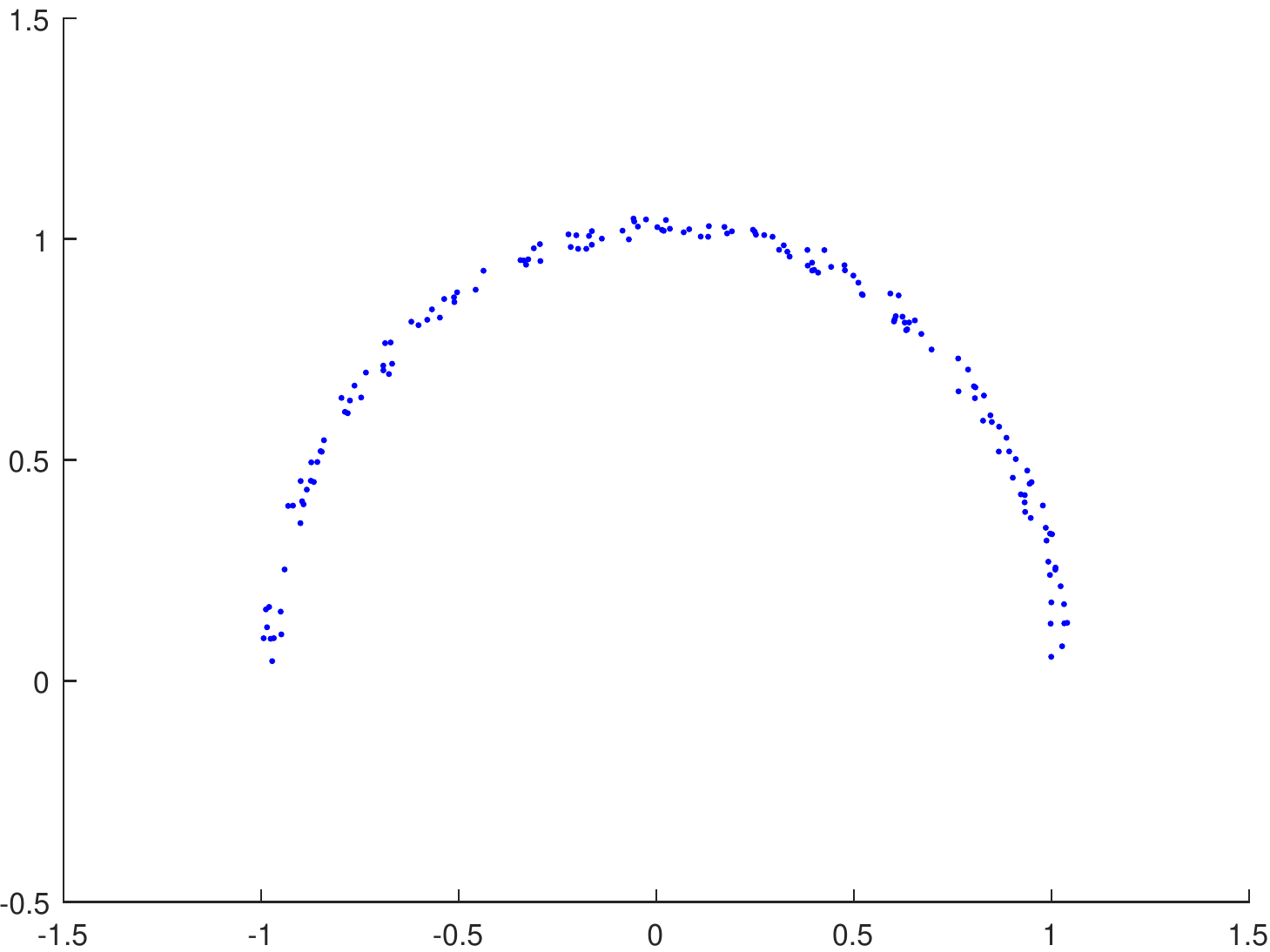}
        \caption{HC dataset}
        \label{fig:halfcircle}
    \end{subfigure}~
    \begin{subfigure}[t]{0.4\textwidth}
        \includegraphics[width=\textwidth]{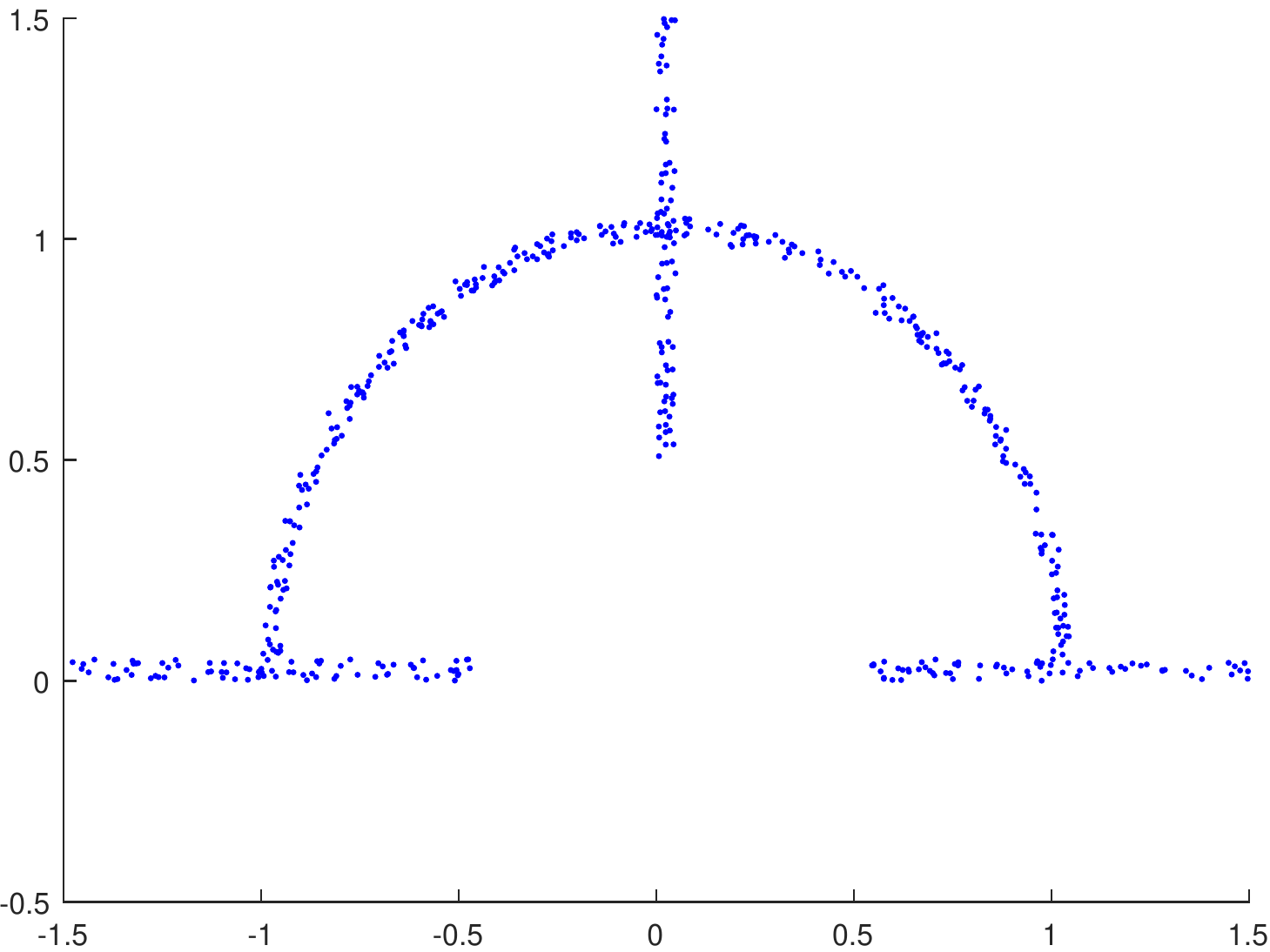}
        \caption{HCW dataset}
        \label{fig:halfcirclewedges}
    \end{subfigure}
    \caption{HC dataset and HCW dataset}
\end{figure}

\begin{figure}[t!]
\centering
    \begin{subfigure}[t]{0.4\textwidth}
        \includegraphics[width=\textwidth]{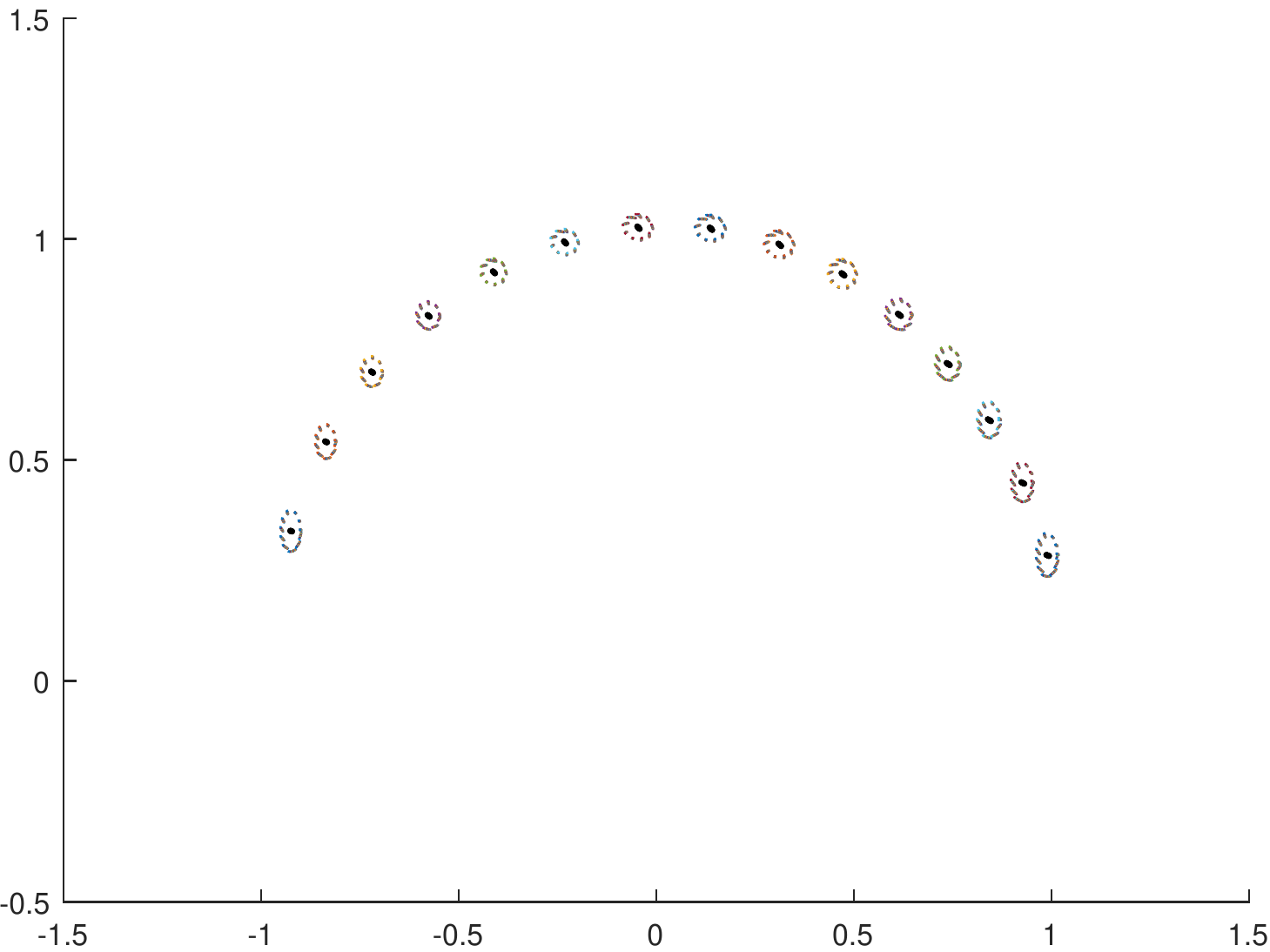}
        \caption{VAE on HC }
        \label{fig:halfcircle_manifold_VAE}
    \end{subfigure}~
    \begin{subfigure}[t]{0.4\textwidth}
        \includegraphics[width=\textwidth]{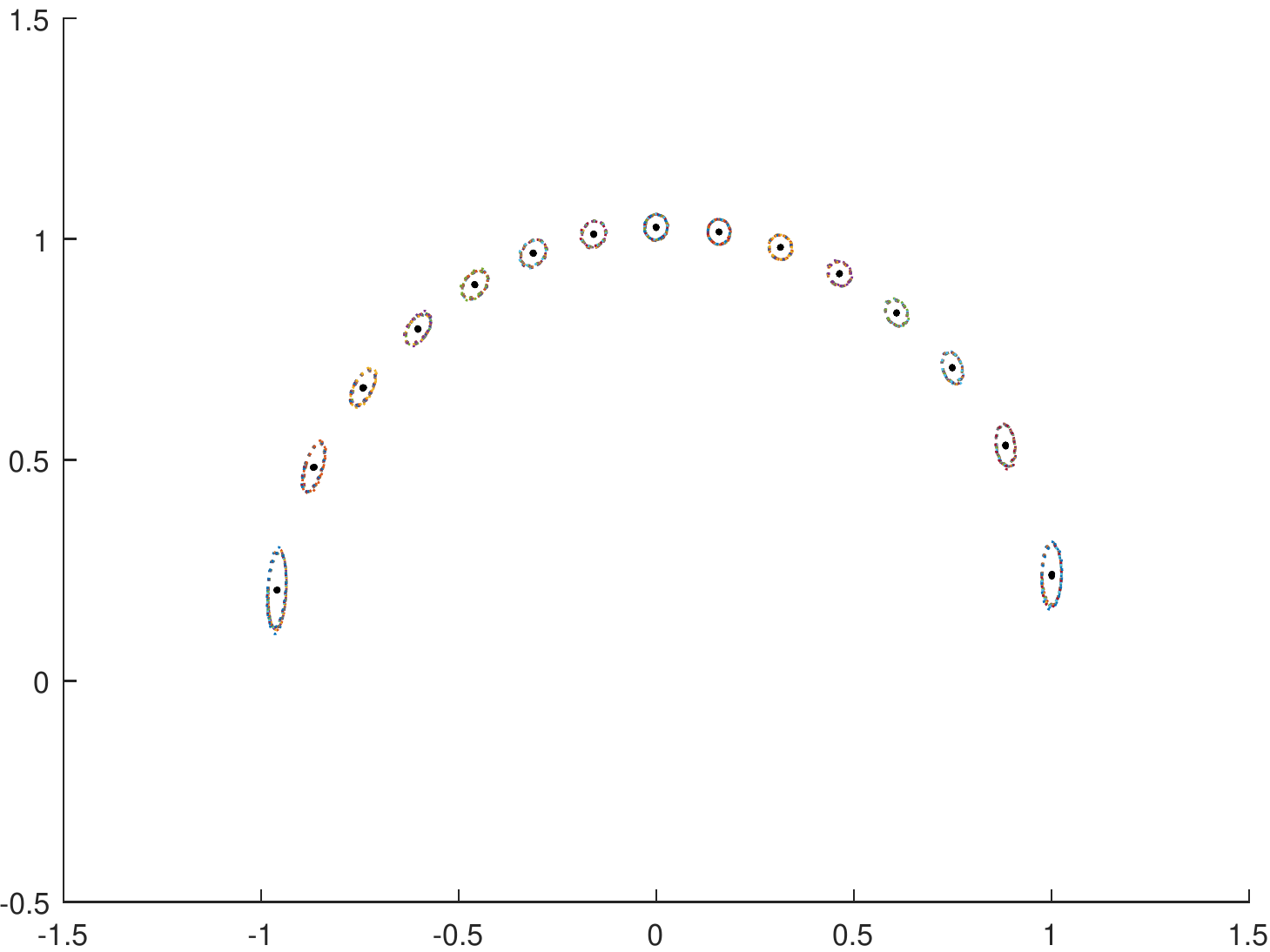}
        \caption{VAE-ROC on HC}
        \label{fig:halfcircle_manifold_VAE-ROC}
    \end{subfigure}\\
    \begin{subfigure}[t]{0.4\textwidth}
        \includegraphics[width=\textwidth]{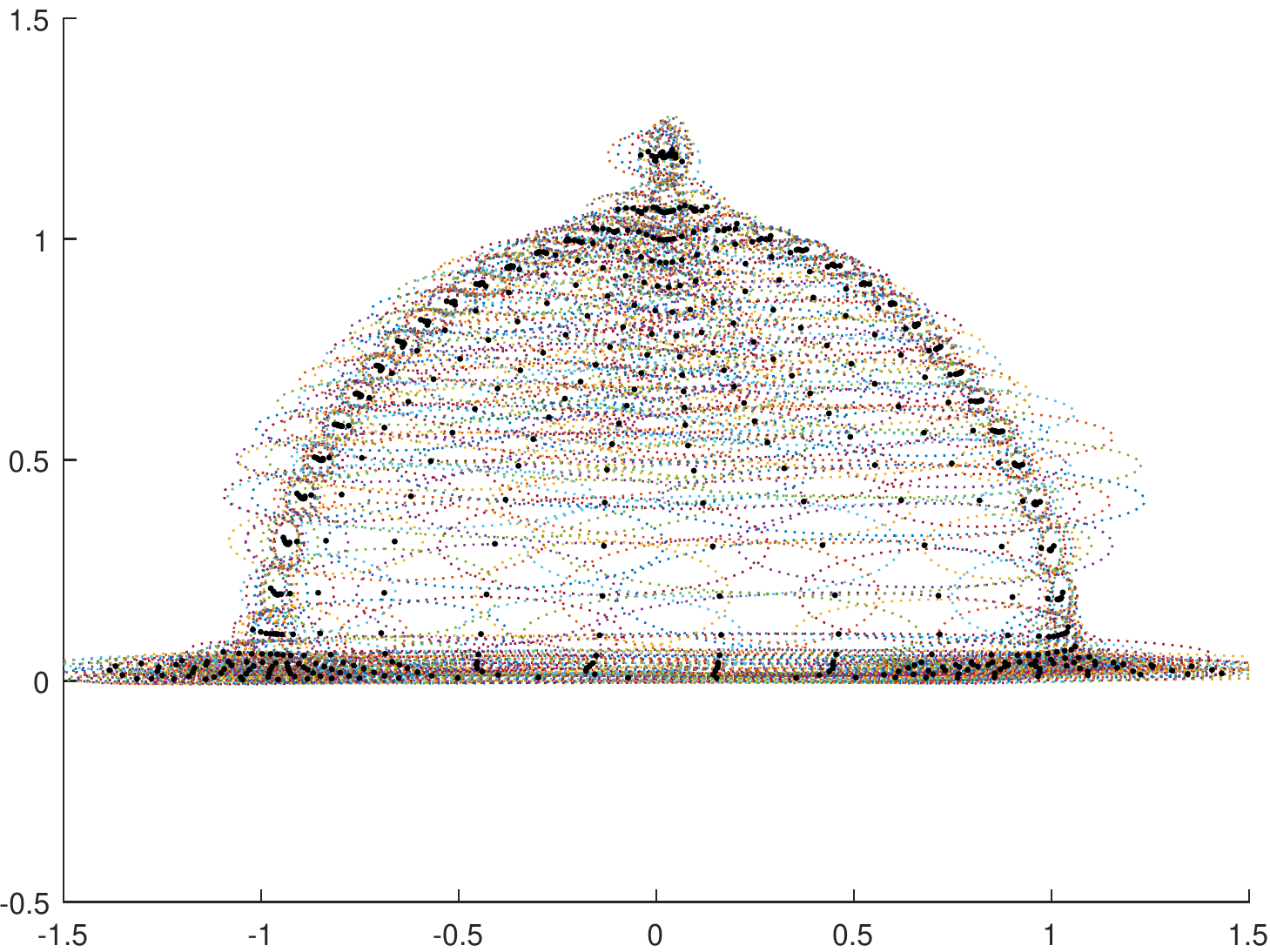}
        \caption{VAE on HCW }
        \label{fig:halfcirclewedges_manifold_VAE}
    \end{subfigure}~
    \begin{subfigure}[t]{0.4\textwidth}
        \includegraphics[width=\textwidth]{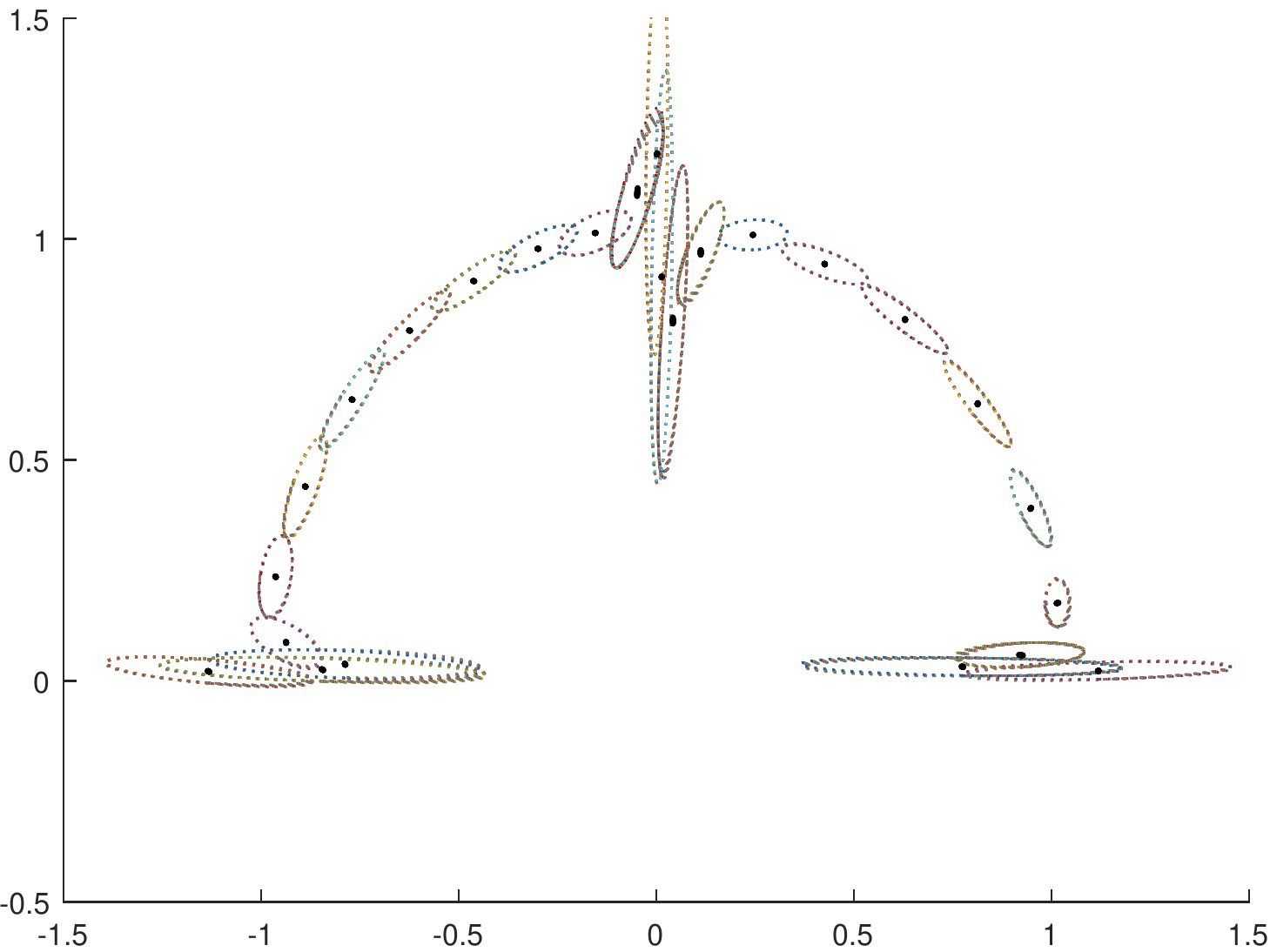}
        \caption{VAE-ROC on HCW}
        \label{fig:halfcirclewedges_manifold_VAE-ROC}
    \end{subfigure}
    \caption{Manifold captured by VAE and VAE-ROC on HC and HCW dataset}\label{fig:halfcirclewedges_manifold}
\textbf{Top row:} the captured manifold on HC dataset with $15 \times 15$ samples. \textbf{Bottom row:} the captured manifold on HCW dataset with $25 \times 25$ samples. A black dot represents a sample from the latent distribution and the dotted contour around the point denotes 3 sigma deviation from the mean(the black dot).
\end{figure}

\begin {table}[t]
\begin{center}
\caption {Log-likelihood of test data on half circle, half circle with wedges and Frey faces dataset. 10,000 instances are sampled to fit Gaussian Parzen window.
We select the best smoothing free parameter with validation set. The value is higher, the better the model captures the true manifold.}
\label{tab:ll_VAE-ROC}
\begin{tabular}{ c ||c | c | c}
    & HC (10K) & HCW (10K)  & Frey Faces (10K)\\
  \hline \hline
  VAE & $1.250 \pm 0.02$ & $0.293 \pm 0.05$    & $722.435 \pm 0.93$\\
  VAE-ROC & $\mathbf{1.431 \pm 0.01}$ & $\mathbf{0.416 \pm 0.02}$ &  $\mathbf{733.553 \pm 1.85}$\\
  VAE-ROC (RR) & - & - & $\mathbf{774.379 \pm 0.79}$\\
\end{tabular}
\end{center}
\end {table}

In this subsection, we model \textit{Half Circle dataset} (HC), which contains 280 two dimensional data points for training set as shown in Fig. \ref {fig:halfcircle}, 60 instances for validation set and 60 instances for test set. We finds that both VAE-ROC and VAE are sensitive to the initial parameters. If the initial parameters setting is out of luck, it ends up with local optima. For this reason, we make 50 runs for each model with random initial parameters and take the best performing model in term of test log-likelihood. Both models has the same number of hidden units (100) and latent dimensions (2). Because the number training of instances is just 260, we use the entire dataset instead of adopting random mini-batch in this dataset.

Fig. \ref{fig:halfcircle_manifold_VAE} and \ref{fig:halfcircle_manifold_VAE-ROC} show the manifold VAE and VAE-ROC capture, which are demonstrated by sampling equally distributed points on the range of cumulative distribution function of multivariate normal distribution (the prior distribution $p(\bz)$). We draw 15 $\times$ 15 samples from the prior distribution. With these coarsely distributed samples in Fig. \ref{fig:halfcircle_manifold_VAE} and Fig. \ref{fig:halfcircle_manifold_VAE-ROC}, we can check how the shape of covariance matrix of each model on the manifold clearly. While the covariance matrix of VAE has just three types of contour (isotropic, fat and tall), that of VAE-ROC can have various tiled forms thanks to the outer product of principal direction, which is clearly helpful to capture true manifold in form of mixture of Gaussian pancakes.

To show usefulness of VAE-ROC on dataset with more complex manifold, we add three wedges (each wedge is consist of 100 instances) to the smooth manifold on  HC in Fig.\ref{fig:halfcirclewedges} and refer it as \textit{Half Cirlce with Wedges dataset} (HCW).
Although smooth manifold in HC is correctly captured by VAE as well as VAE-ROC, in case of HCW, it is hard to get the right manifold with VAE even we use ADAM, which is a adaptive learning rate algorithm that can cope with local optima effectively. After extensive search of hyper-parameters with validation set, VAE and VAE-ROC capture the manifold as shown in Fig. \ref{fig:halfcirclewedges_manifold_VAE} and Fig. \ref{fig:halfcirclewedges_manifold_VAE-ROC} respectively. While some of the density seep out of the true manifold in VAE, VAE-ROC models the complex manifold by using two components; smooth manifold of half circle is capture with $\bmu$ and complex manifold represented by three wedges is modeled by principal direction vector $\ba$ in Gaussian output. Quantitatively, per instance test log-likelihood measured by Gaussian Parzen window fit with the samples generated by both models shows that VAE-ROC is closed to the true data-generating distribution rather than VAE as shown in Table. \ref{tab:ll_VAE-ROC}.

\begin{figure}[t!]
    \centering
    \begin{subfigure}[t]{0.4\textwidth}
        \includegraphics[width=\textwidth]{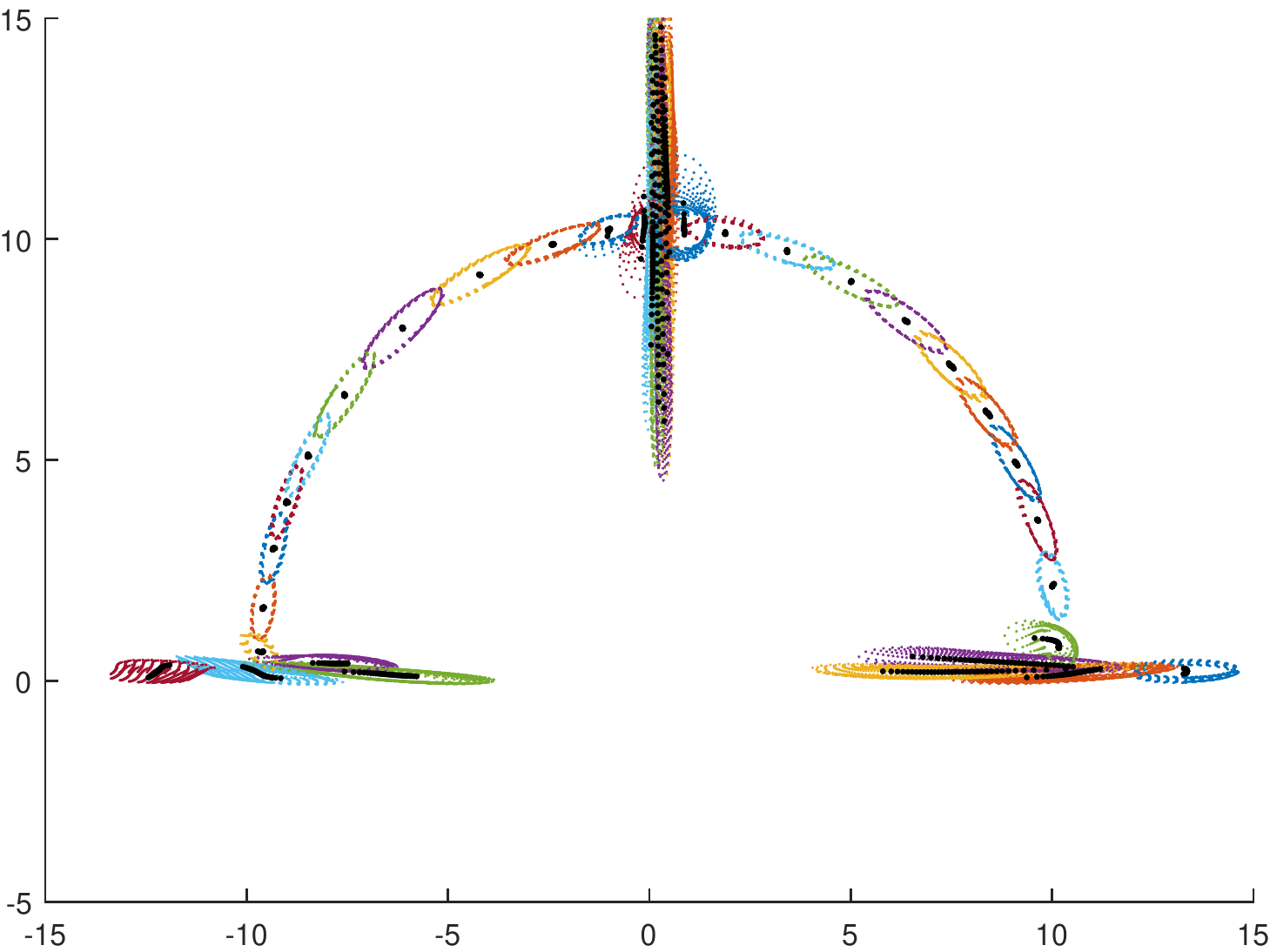}
        \caption{Manifold capture by VAE-ROC with L2 norm regularization $\lambda^{local}=0.1$}
        \label{fig:halfcirclewedges_reg1}
    \end{subfigure} ~
    \begin{subfigure}[t]{0.4\textwidth}
        \includegraphics[width=\textwidth]{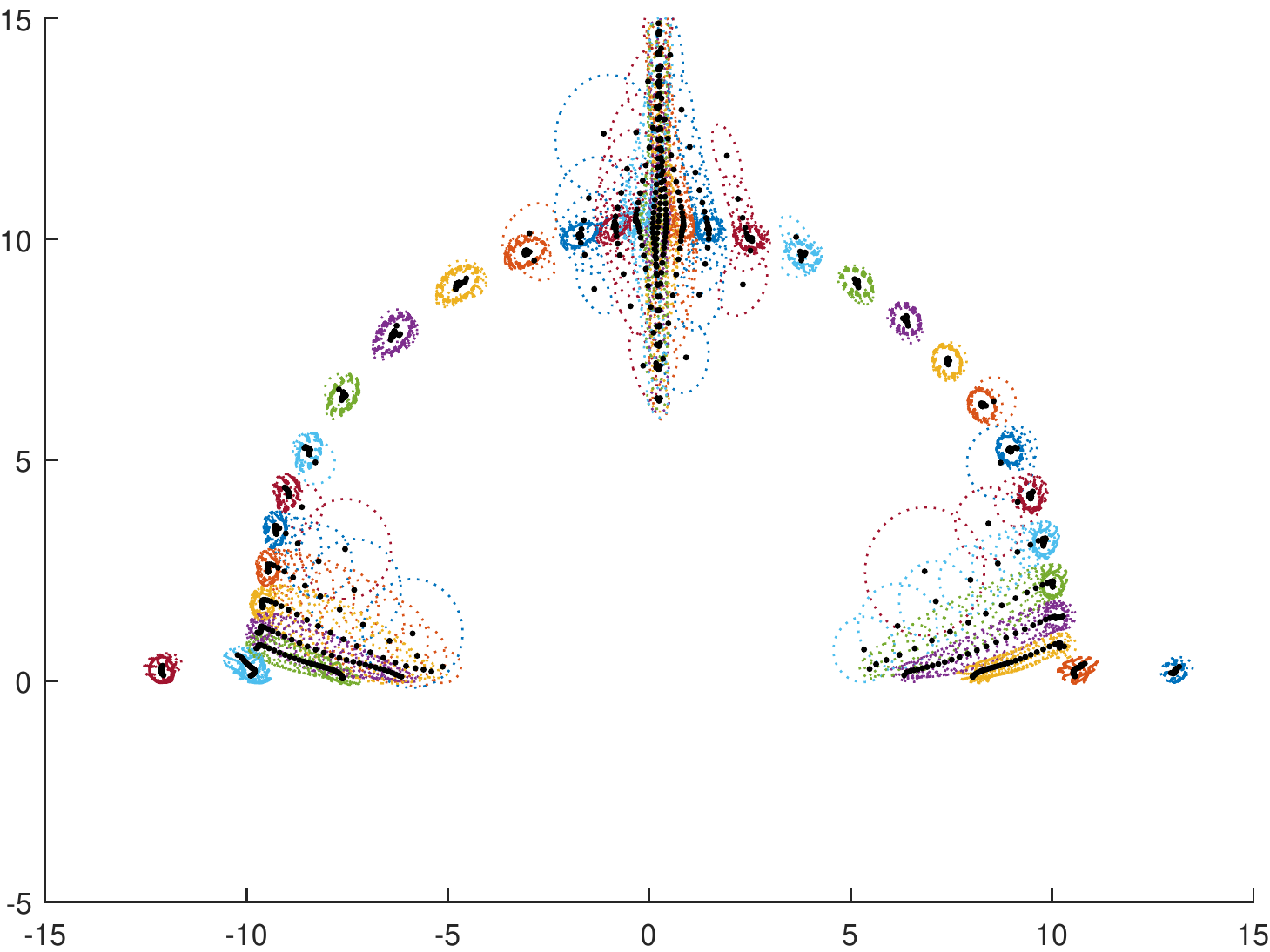}
        \caption{Manifold capture by VAE-ROC with L2 norm regularization $\lambda^{local}=5$}
        \label{fig:halfcirclewedges_reg3}
    \end{subfigure}
    \caption{Captured manifold by VAE-ROC with the different level of L2 norm regularization on HCW}\label{fig:halfcirclewedges_reg}
\end{figure}

Now, let us delve into regularization and to find how much it affect the result of learning with VAE-ROC. Actually, it is hard to learn the true manifold with VAE on HCW dataset because the restriction on the types of Gaussian contour hamper the model to learn the true manifold. However, VAE-ROC without norm regularization, learn the true manifold easily but some data only exist on principal direction vector not on the  smooth manifold captured by $\bmu$. Due to this reason, some data is not reconstructed well enough in a perspective of autoencoder. To cope with this problem, we lay L2 regularization and the results are reported in Fig. \ref{fig:halfcirclewedges_reg}. The more we weigh L2 regularization term, the width of Gaussian contour of each mixure component shrinks more.
Fig. \ref{fig:halfcirclewedges_reg3} shows similar width of Gaussian contour with Fig. \ref{fig:halfcirclewedges_manifold_VAE} but it captures true manifold much more correctly than VAE. This characteristic of regularized VAE-ROC enable us to learn true manifold of MNIST in later experiment.

\subsection{Frey Faces}

The Frey face dataset is a set of face images where we randomly divide total 1,965 data points into three subsets. As a result, we have 1,573 points for the training set, 196 points for the test set and 196 points for the  validation set. We learn VAE and VAE-ROC on the training set and select L2 regularization parameter $\lambda^{local}$ with the validation set. Considering the range of value [0,1], we adopt the sigmoid function for each mean parameter of Gaussian output following \cite{KingmaDP2014iclr}.  Both models have 2-D latent space with two layers, each of which has 100  ReLU type hidden units.

\begin{figure}[hb!]
    \centering
        \includegraphics[width=0.6\textwidth]{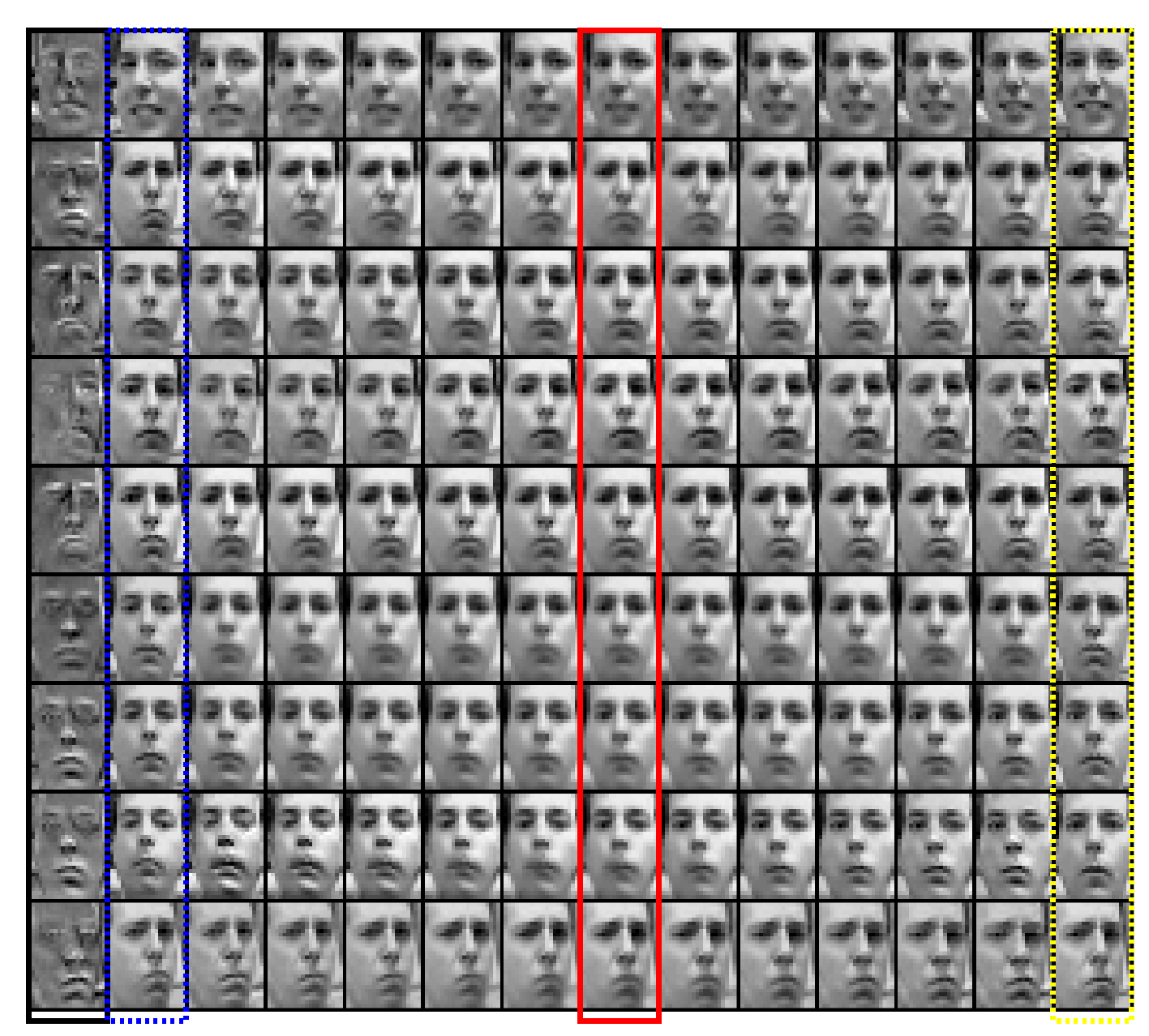}
    \caption{ Tangent spaces on the smooth manifold captured by VAE-ROC. The each row denotes a tangent space on the smooth manifold centered on $\bmu$ (red solid box).   Each image of the first column denotes the tangent vector for each row. From the images on the smooth manifold, a tangent space is spanning with the corresponding $\ba$. Images are sampled from the corresponding Gaussian pancake(tangent space) with equally distributed points on the range of cumulative distribution function of Gaussian distribution. The bue and yellow colored dashed boxes contain the most closest images in the training set to the nearest sampled images}\label{fig:ff_tanget_spaces_VAE-ROC}
\end{figure}

 As shown in Fig. \ref{fig:ff_tanget_spaces_VAE-ROC}, VAE-ROC captures tangent space that is centered on the smooth manifold, which is modeled by the mean face $\bmu$. The rest of manifold deviated from the smooth manifold is captured by the principal direction vector $\ba$. The larger the norm of $\ba$ is, the larger deviation from the smooth manifold can be captured. By regularizing the norm of $\ba$, we can control how much proportion of manifold is captured by the smooth manifold. With cross validation, we find $\lambda^{local} = 100$ works best. Looking closely into the tangent space in the Fig.   \ref{fig:ff_tanget_spaces_VAE-ROC}, each row shows faces on the tangent space of the mean face (red box at the center) that is deviated with the different weighted principal direction vectors (black box at the leftmost column).

We also confirm effectiveness of VAE-ROC quantitatively with the test log-likelihood of both models measured by Gaussian Parzen window as shown in Table. \ref{tab:ll_VAE-ROC}, where Rank Regularization (RR) is also applied to improve the quality of samples that VAE-ROC generates.


\subsection{MNIST Dataset}
\begin{figure}[ht!]
    \centering
    \begin{subfigure}[t]{0.4\textwidth}
        \includegraphics[width=\textwidth]{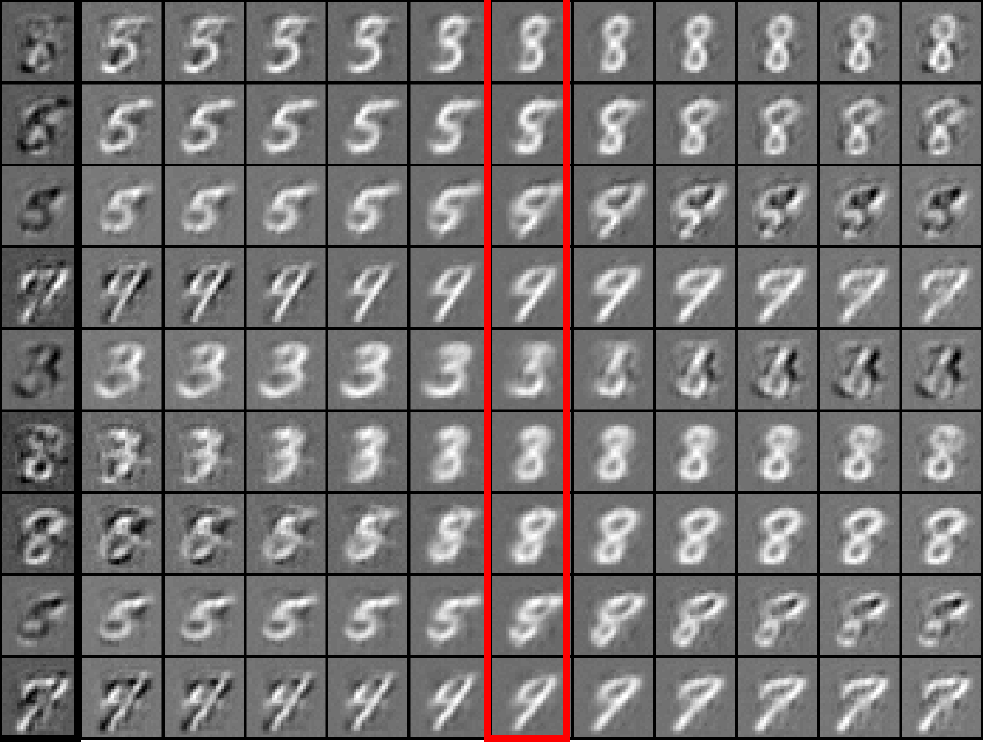}
        \caption{Randomly sampled 9 tangent spaces with VAE-ROC. ($\lambda^{local}=100$ with L1 regularization)}
        \label{fig:tanget_spaces_VAE-ROC}
    \end{subfigure}
    \begin{subfigure}[t]{0.4\textwidth}
        \includegraphics[width=\textwidth]{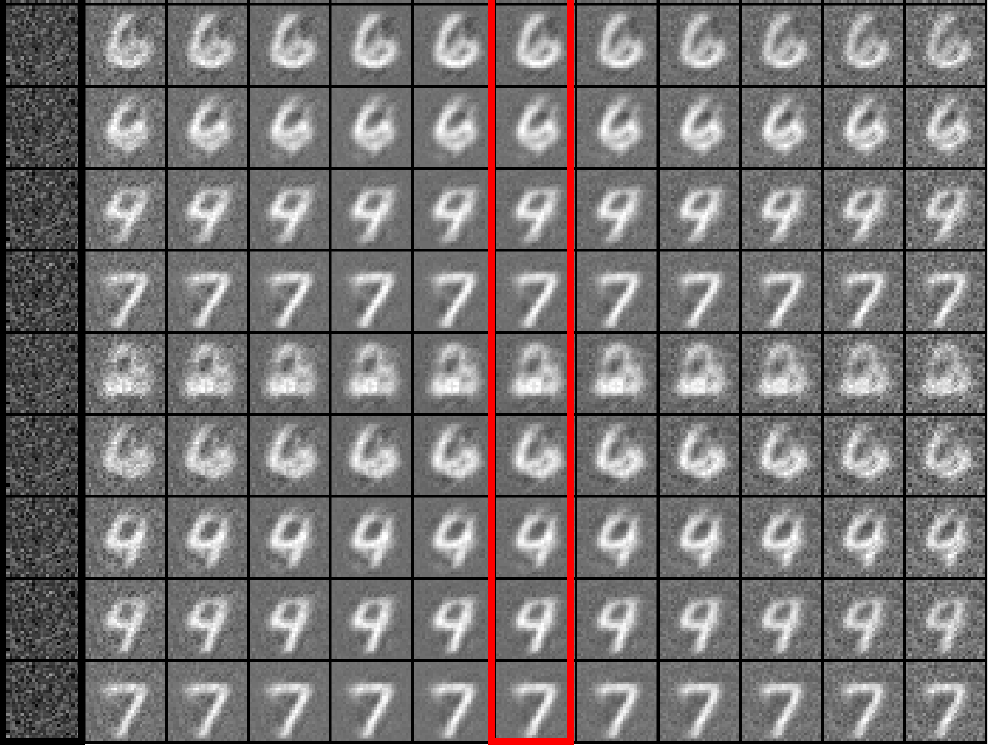}
        \caption{Randomly sampled 9 tangent spaces with VAE-ROC. ($\lambda^{local}=1000$ with L1 regularization)}
        \label{fig:tanget_spaces_VAE-ROC_1000}
    \end{subfigure}
    \caption{A manifold captured by VAE-ROC on the MNIST dataset with tangent spaces on it.
    The first column in each panel denotes the $\ba$ that chracterizes principal direction vector of the tangent space and the image of the center in the red box denotes the image $\bmu$ on the manifold.}
    \label{fig:MNIST_VAE-ROC}
\end{figure}

\begin {table}[ht!]
\begin{center}
\caption {The test log-likelihood of other [0,1] real valued generative models. We get the numbers from each paper. Adversarial Auto Encoder (Adv. AE) shows the best but it uses cross entropy error and it improves the performance by regularizing aggregate posterior by GAN. Thus, we can use Adv. AE with VAE-ROC.}
\label{tab:ll_MNIST_VAE-ROC}
\begin{tabular}{ l ||c }
    & MNIST (10K) \\
  \hline \hline
  DBN \cite{HintonGE2006neco} & $ 138 \pm 2$ \\
  GAN \cite{GoodfellowIJ2014nips} & $ 225 \pm 2$ \\
  GMMN + AE \cite{LiY2015icml} & $ 282 \pm 2$ \\
  VAE-ROC & $\mathbf{292 \pm 1}$ \\
  VAE-DGF(5-500-300)\cite{DaiZ2016iclr} & $301.67$  \\
  VAE-ROC(RR) & $\mathbf{303 \pm 0.76}$ \\
  Adv. AE \cite{MakhzaniA2016iclr} & $ 340 \pm 2$ \\
\end{tabular}
\end{center}
\end {table}
In this subsection, we learn VAE-ROC on real valued MNIST dataset \cite{LeCunY98procieee} as shown in Fig. \ref{fig:MNIST_VAE-ROC}, where we only report the results of VAE-ROC because we failed to get the comparable result using Gaussian VAE with diagonal covariance matrix. Even with intensive hyper parameter search with random initial parameters, it is very hard to train the Gaussian VAE with diagonal covariance matrix on real valued MNIST dataset. However, by regularizing VAE-ROC with extreme weighted L1 regularization, VAE-ROC learns the manifold that Gaussian VAE was supposed to learn as shown in Fig. \ref{fig:tanget_spaces_VAE-ROC_1000}, where we are only to find noise in the tangent vector, which means its covariance matrix is closed to diagonal matrix.

 To verify VAE-ROC quantitatively with existing models, we randomly divide the given 60,000 training set into 50,000 training set and  10,000 validation set in this experiment. We also scale the value into [0,1] to compare with other models. Moreover, we optimize our model with random hyper parameter search with 100 iterations and compare the test log-likelihood measured by Gaussian Parzen windows (fitted by 10,000 samples) where we find free parameter $\sigma$ with validation set. We report the result in Table \ref{tab:ll_MNIST_VAE-ROC}.
\begin{figure}[th!]
    \centering
    \begin{subfigure}[t]{0.4\textwidth}
        \includegraphics[width=\textwidth]{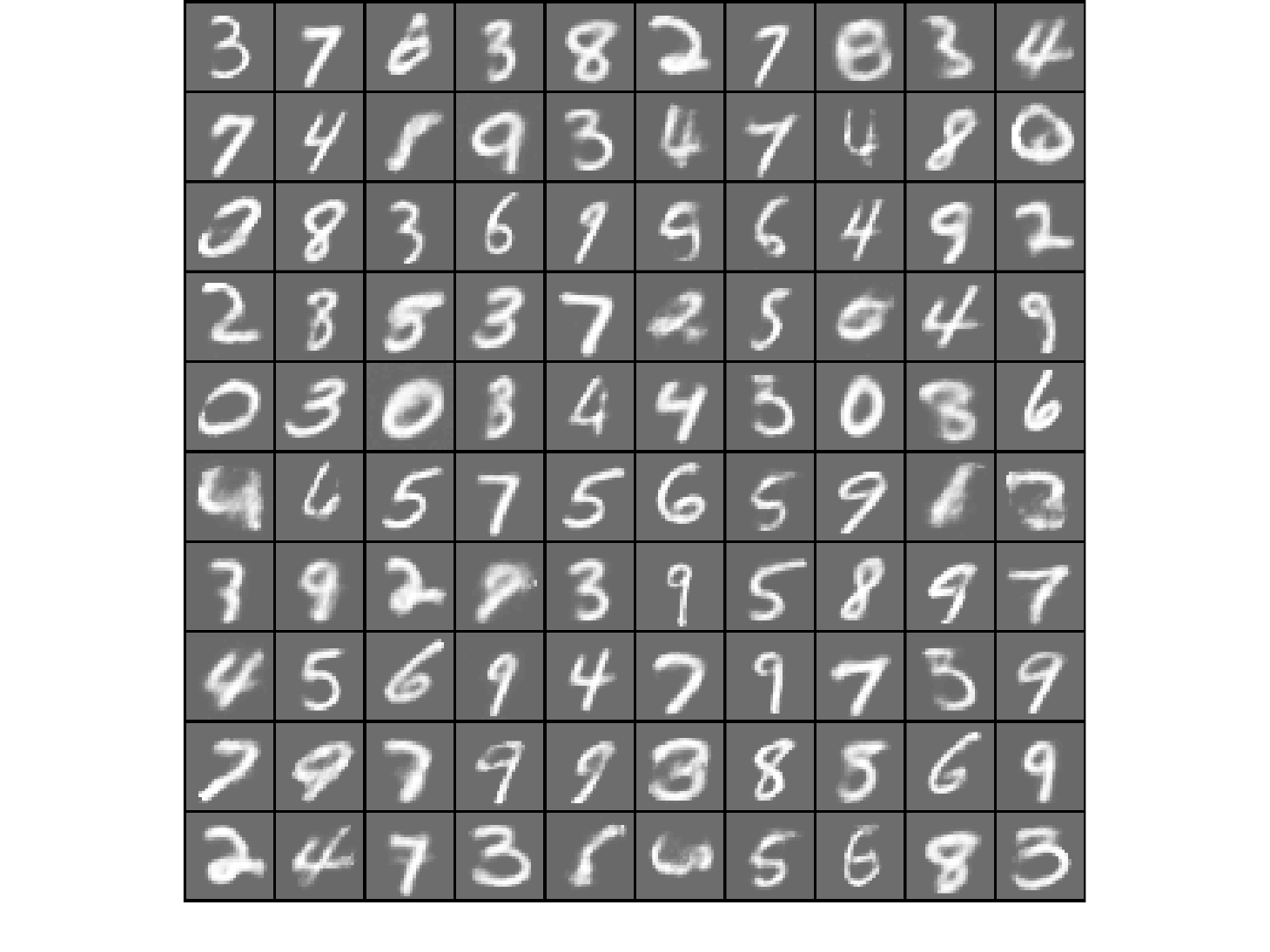}
        \caption{100 samples from VAE-ROC(RR)}
        \label{fig:mnist_best_samples}
    \end{subfigure}
    \begin{subfigure}[t]{0.4\textwidth}
        \includegraphics[width=\textwidth]{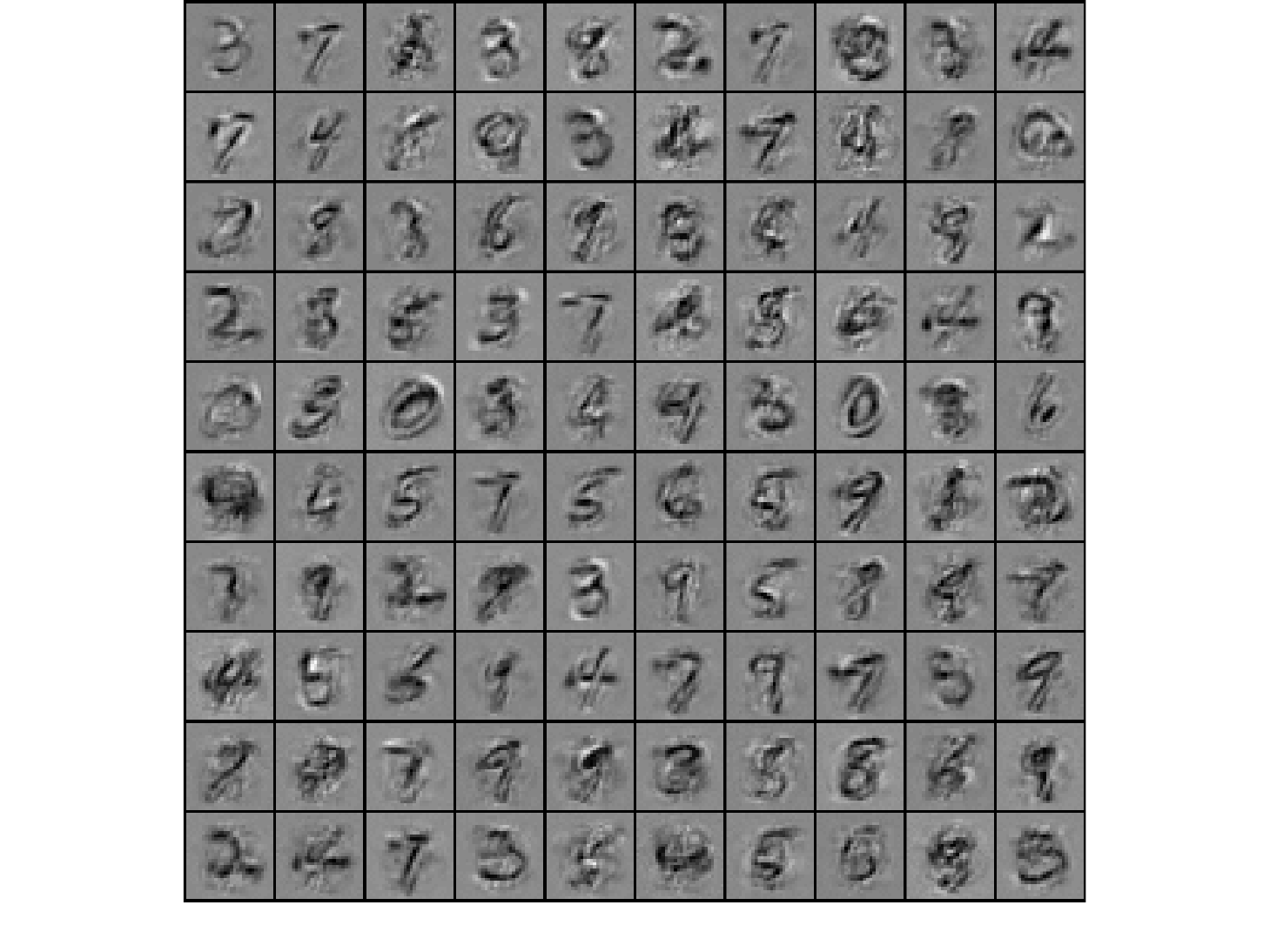}
        \caption{ The corresponding principal direction vectors for the samples}
        \label{fig:mnist_best_pdv}
    \end{subfigure}\\
    \begin{subfigure}[t]{0.4\textwidth}
        \includegraphics[width=\textwidth]{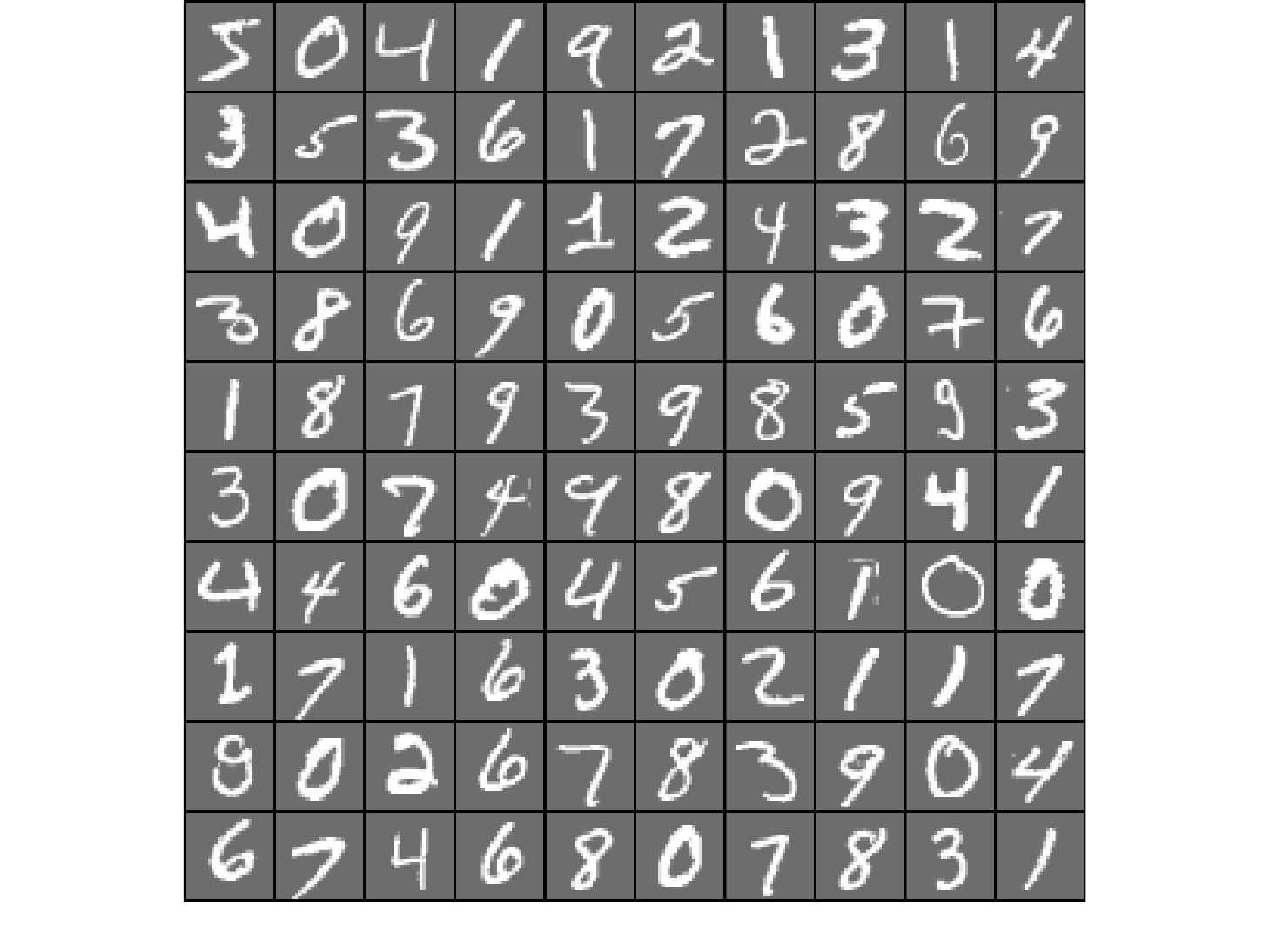}
        \caption{MNIST dataset.}
        \label{fig:mnist_best_data}
    \end{subfigure}
    \begin{subfigure}[t]{0.4\textwidth}
        \includegraphics[width=\textwidth]{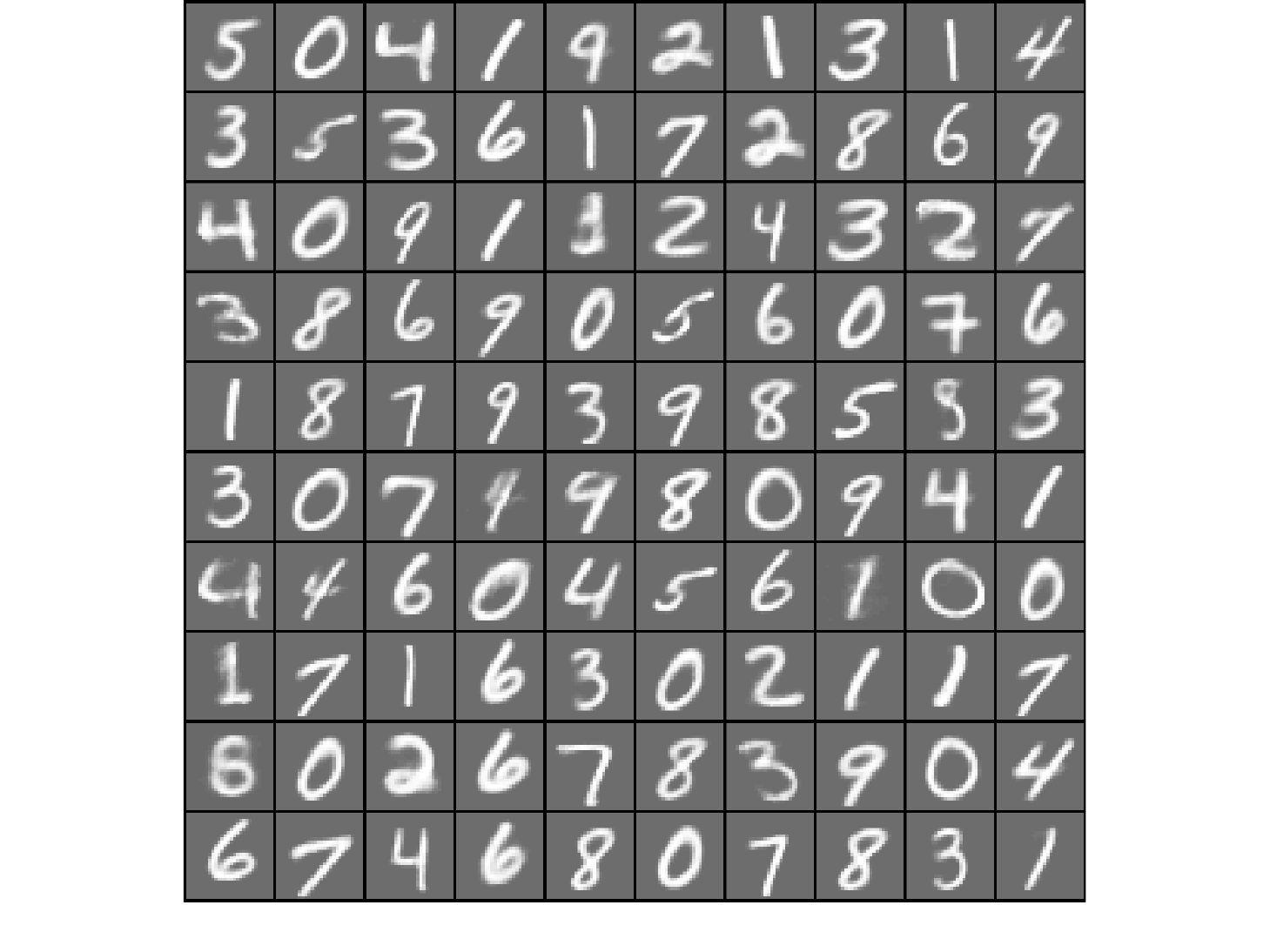}
        \caption{Reconstructed dataset by VAE-ROC(RR)}
        \label{fig:mnist_best_recon_data}
    \end{subfigure}
    \caption{VAE-ROC(RR) on MNIST dataset (\textbf{VAE-ROC:} 5 latent dimensions, 767 ReLU hidden units for two layers, L2 regularization with $\lambda^{local} = 0.1$. \textbf{VAE-ROC(RR):} 5 latent dimensions, 635 ReLU hidden units for two layers, L2 regularization with $\lambda^{local} = 10$.}\label{fig:mnist_best}
\end{figure}

\begin{figure}[hb!]
    \centering
        \includegraphics[width=0.40\textwidth]{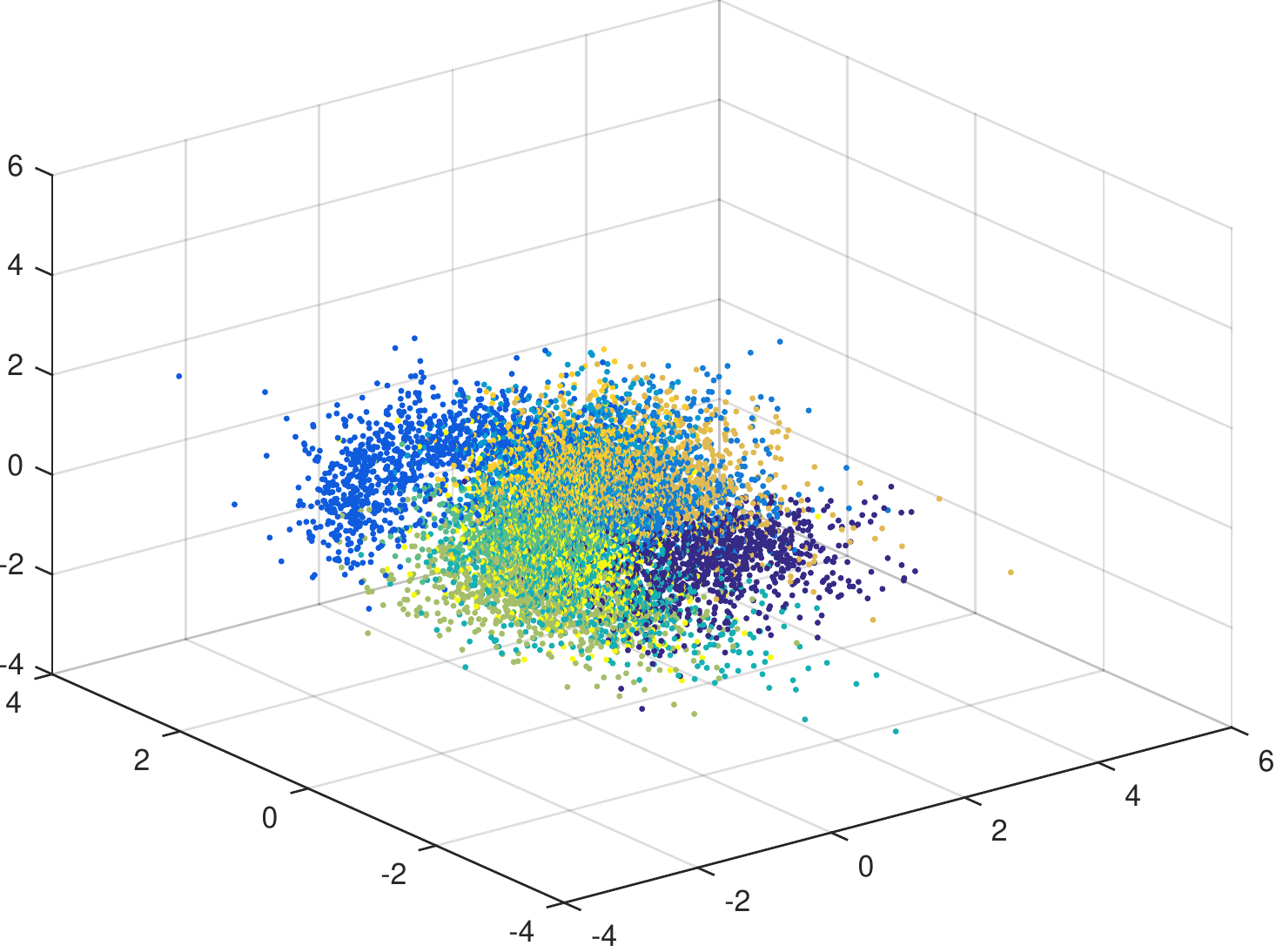}
        \caption{3-D latent z code on validation dataset. we choose three dimesions for visualization.}
        \label{fig:minst_best_zcode3d}
\end{figure}

Qualitatively, we compare the generated samples and reconstructed images from VAE-ROC (RR) with the original images in Fig. \ref{fig:mnist_best}. As shown Fig. \ref{fig:mnist_best_recon_data}, VAE-ROC (RR) successfully reconstructs the given data, which means MLP learns nonlinear basis for MNIST dataset. Based on these nonlinear basis, VAE-ROC (RR) generates digits with various style as shown in Fig. \ref{fig:mnist_best_samples} with the corresponding principal direction vector $\ba$ in Fig. \ref{fig:mnist_best_pdv}. Moreover, we verify the shape of latent code $\bz$ on validation dataset in Fig. \ref{fig:minst_best_zcode3d}. It shows isotropic contour, which  enables VAE-ROC (RR) to generate quality samples than VAE-ROC.
\subsection{UCI Auto Dataset}
In the following two subsections, we compare GCVAE with VAE on two mixed variate datasets. In this subsection, we first visualize UCI Auto dataset with one hot encoding categorical variables and normalized continuous variables with zero mean and unit variance, which let us check whether the model is actually trained in SGD.  UCI Auto dataset consists of 392\footnote{Actually, it contains 398 instances, six instances of which have missing entry. We omit these instances with missing entry. Due to the lack of data, we use only 30 hidden units and 2 dimensional latent space and early stopping at 10,000 iterations of parameter updates.} instances of automobiles, each of which has 8 dimensional values.  Five of them are continuous variables and the rest of them are categorical variables (i.e. Country of origin, the number of cylinders and the model year). We randomly select 313 instances as a training set, 39 instances as validation set and the rest 40 instances as test set.  We train GCVAE described in the subsection. \ref{subsec:GCoupulaVAE} and the corresponding VAE (Gaussian and categorical outputs without copula) on UCI auto dataset.
\begin{figure}[t!]
    \centering
    \begin{subfigure}[t]{0.5\textwidth}
        \includegraphics[width=\textwidth]{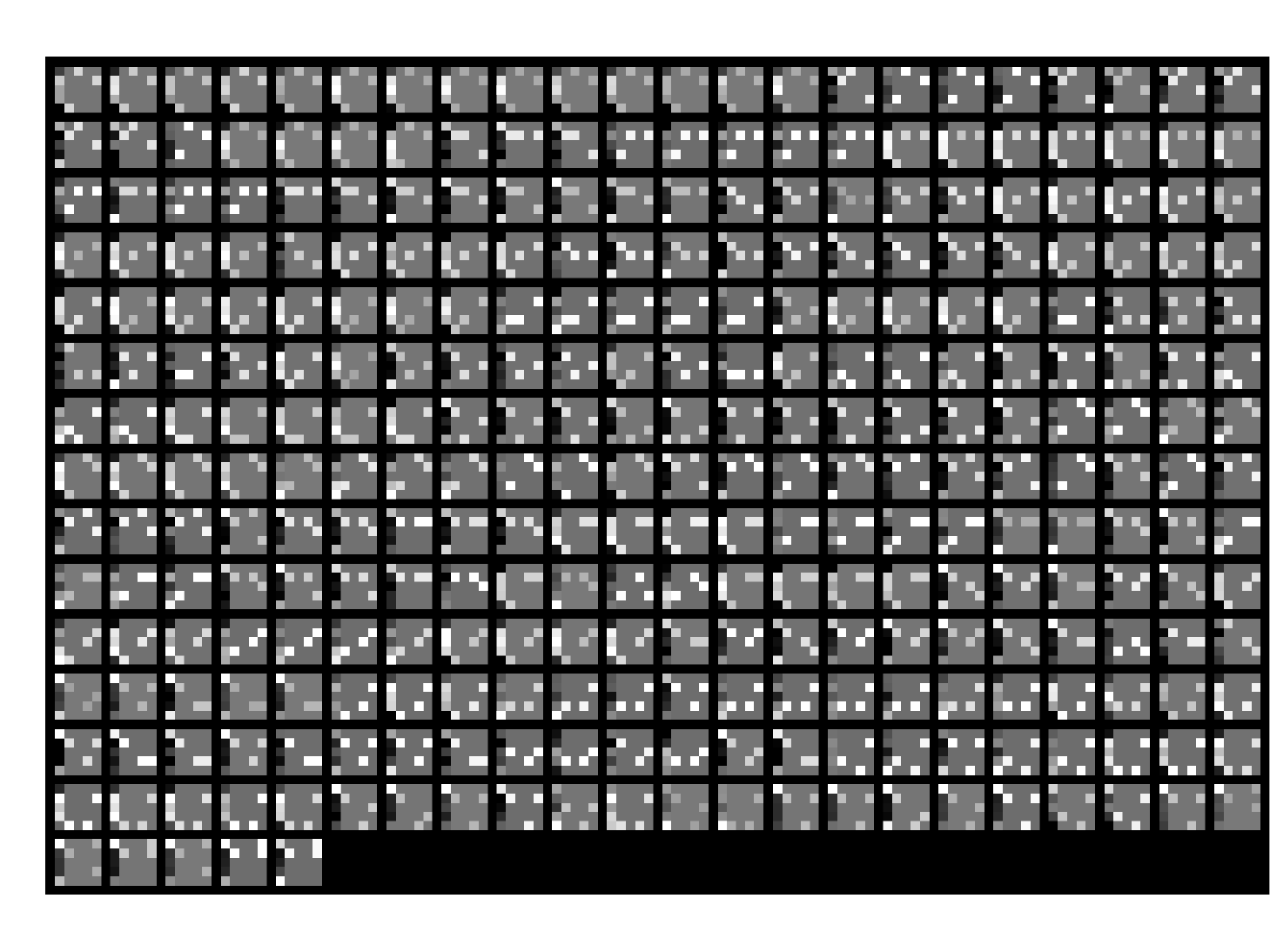}
        \caption{UCI Auto dataset}
        \label{fig:auto_dataset}
    \end{subfigure}\\
    ~
    \begin{subfigure}[t]{0.4\textwidth}
        \includegraphics[width=\textwidth]{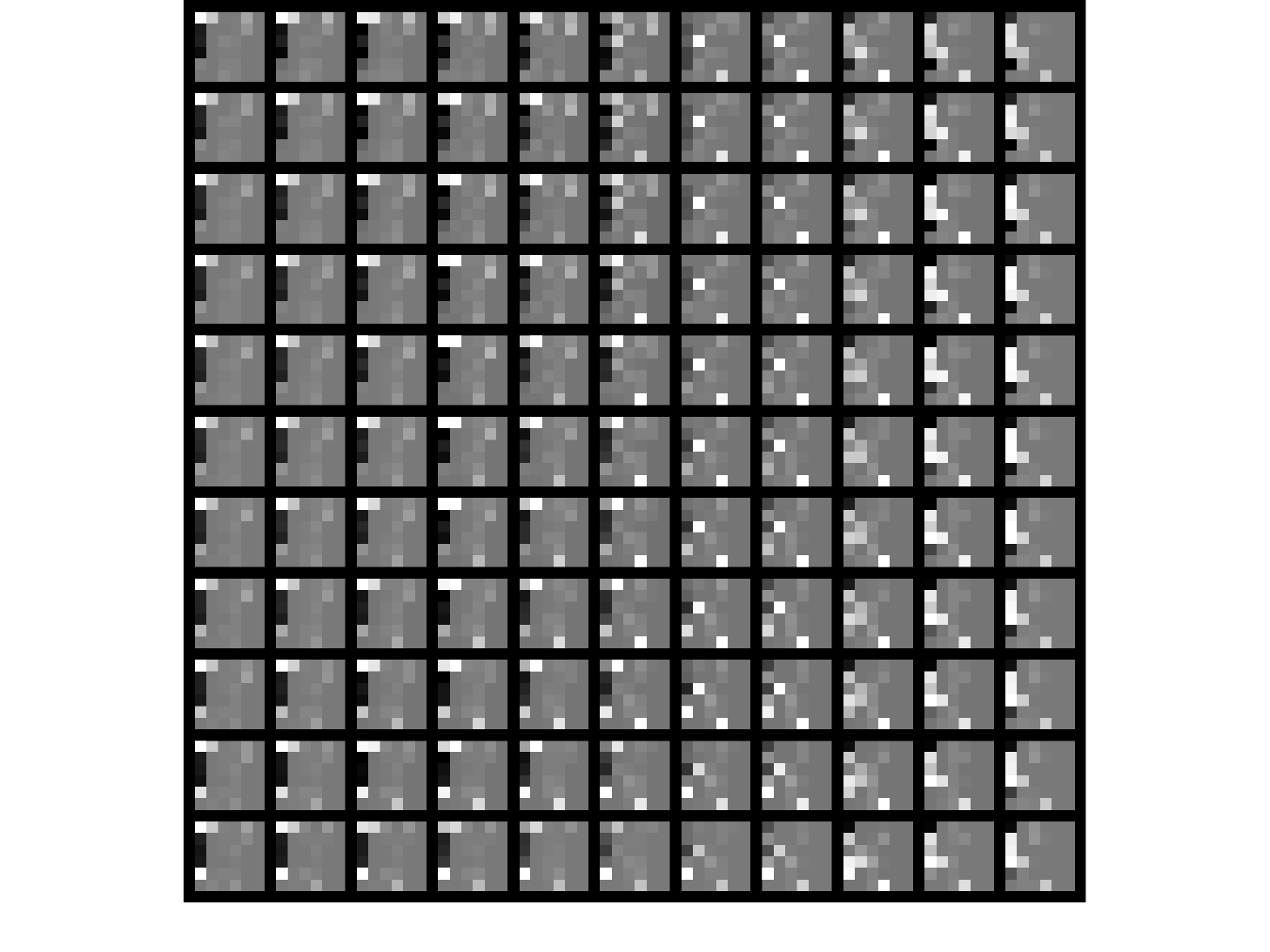}
        \caption{Manifold captured by VAE with 10 $\times$ 10 samples}
        \label{fig:auto_VAE_manifold}
    \end{subfigure}
    ~ 
    \begin{subfigure}[t]{0.4\textwidth}
        \includegraphics[width=\textwidth]{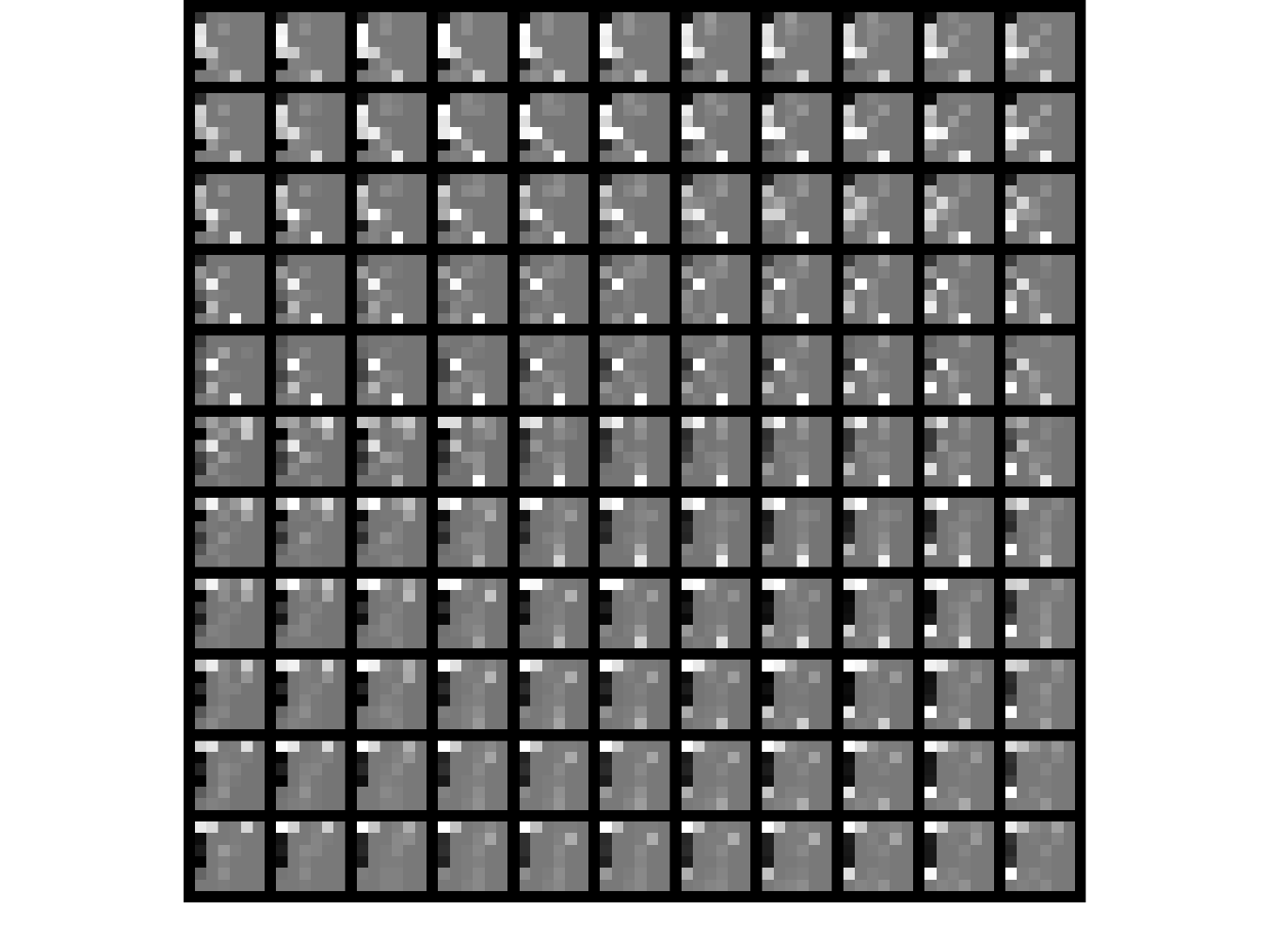}
        \caption{Smooth manifold captured by Gaussian Copula VAE with 10 $\times$ 10 samples}
        \label{fig:auto_Copula_VAE_manifold}
    \end{subfigure}
    \caption{VAE and GCVAE on UCI Auto dataset}\label{fig:uci_auto}
\end{figure}

As a generative model, GCVAE models data generating distribution on UCI Auto dataset, which can be evaluated by comparing the test log-likelihood on the distribution. Instead of using the exact value, we can approximate the distribution with samples from each model, which are used to fit kernel density estimator. In case there are only continuous variables, we may use Gaussian Parzen window method \cite{BengioY2014icml,GoodfellowIJ2014nips}. However, we are treating a mixed variate dataset, which brings the following kernel density estimator for mixed variate dataset ( mixed with continuous and discrete variables ) \cite{AitchisonJ76biometrika}:
{\footnotesize
\bee
\lefteqn{p([\bx^c,\bx^s]) }\\
&&= \frac{1}{N}\sum_{n=1}^{N^{s}}\bigg(\prod_{d=1}^{d_c}\frac{1}{\sigma\sqrt{2\pi}} \exp\big(-\frac{1}{2}\frac{(\bx_d^c -\bx^{c,(n)})^2}{\bsigma^{2}}\big) \prod_{d=1}^{d_s} K(\bx_d^{s},\bx_d^{s,(n)})\bigg)~,
\eee}
where  $N^{(s)}$ is the number of samples to build kernel density estimator (10,000 samples),  $K(\bx_d^{s},\bx_d^{s,(n)})$ denotes a kernel function for categorical variable. It has $1-h$ in case $\bx_d^{s} = \bx_d^{s,(n)}$ and $h/(E-1)$ in otherwise. $h$ and $\bsigma$ are free parameters for kernel density estimator and selected via cross validation\footnote{ $h \in \{ 0.8,0.6,0.4,0.3,0.2,0.1,0.05  \}$,\\$~~~~~~~\sigma \in \{ 1000,100,10,1,0.1,0.01,0.001 \}$. }. We describe how to sample from GCVAE in Appendix. \ref{sec:gc_sample}.

 As shown in Table \ref{tab:ll_copulaVAE}, we take randomly initiated 10 runs for each model and report mean and standard deviation, where we can find GCVAE shows better per-instance test log-likelihood than VAE measured by the kernel density estimator fit by samples from the corresponding model. We use no regularization in this experiment.

\begin {table}[H]
\begin{center}
\caption {Approximated log-likelihood of test data on UCI Auto dataset.} \label{tab:ll_copulaVAE}
\begin{tabular}{ l |c | c }
    & UCI Auto (10K) & UCI SPECT (10K) \\
  \hline
  VAE & $ -200.289 \pm 3.751$ & $ -144.195 \pm 3.443$\\
  GCVAE & $\mathbf{-189.344 \pm 4.599}$ & $ \mathbf{-134.579 \pm 2.621 }$\\
\end{tabular}
\end{center}
\end {table}

\if(0)
\begin{figure*}[t!]
    \centering
    \begin{subfigure}[t]{0.5\textwidth}
        \includegraphics[width=\textwidth]{./Auto_Copula_VAE_ll}
        \caption{Gaussian Copula VAE}
        \label{fig:auto_manifold_CopulaVAE}
    \end{subfigure}
    ~ 
    \begin{subfigure}[t]{0.5\textwidth}
        \includegraphics[width=\textwidth]{./Auto_VAE_ll}
        \caption{VAE}
        \label{fig:auto_manifold_VAE}
    \end{subfigure}
    \caption{Approximate ELOB of CopulaVAE and VAE to model continuous and categorical random variables at the same time. }\label{fig:UCI_auto_result}
\end{figure*}
\fi

\begin{figure}[th]
    \centering
        \includegraphics[width=0.5\textwidth]{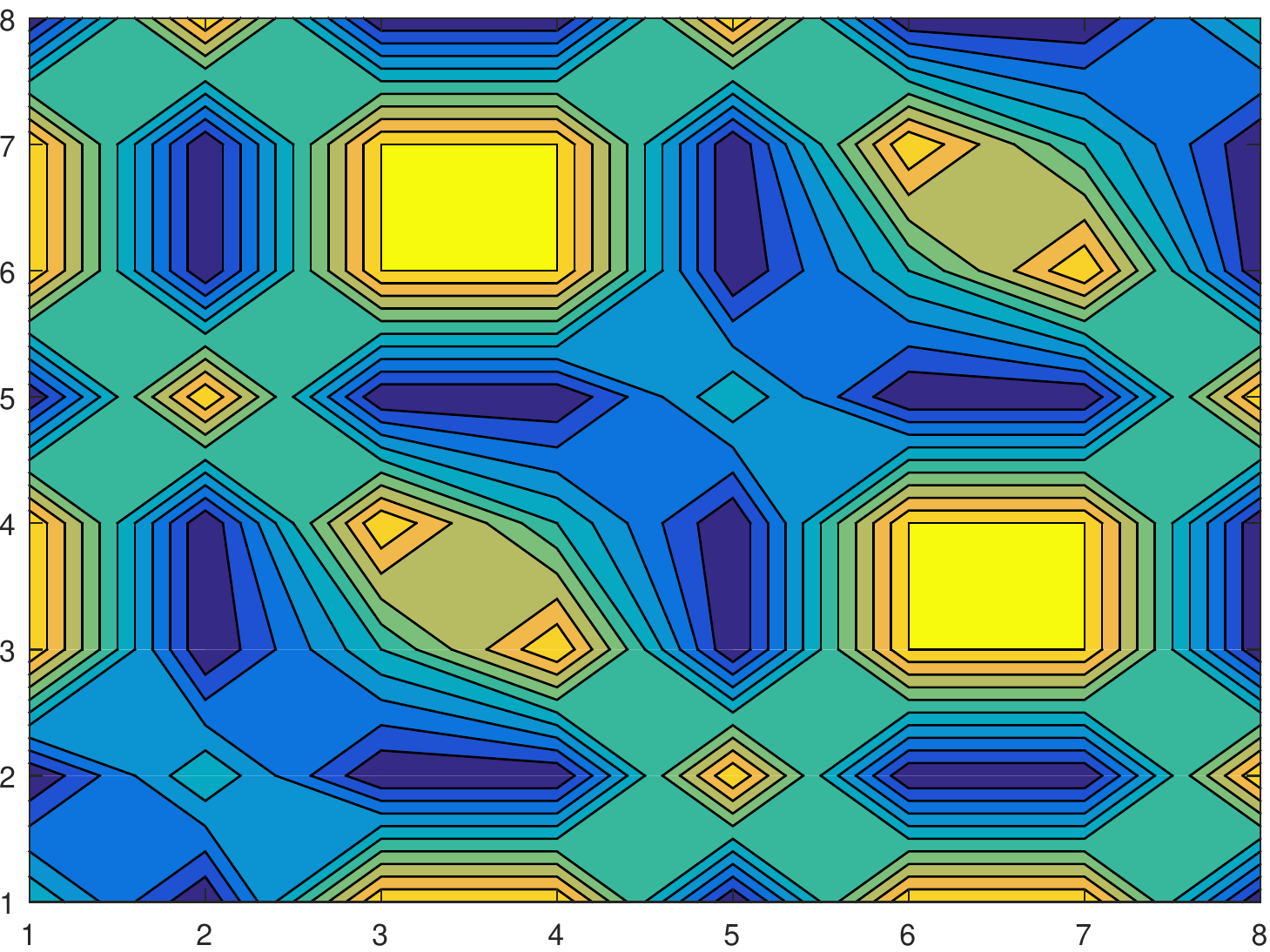}
        \caption{Rank correlation in one sample from Gaussian Copula VAE on UCI Auto dataset. (1: cylinders, 2:model year, 3:origin, 4:mpg, 5:displacement, 6:horsepower, 7:weight, 8:acceleration) }
        \label{fig:rc_uci}
\end{figure}

Qualitatively, we also check local dependency with Kendall's tau as shown Fig. \ref{fig:rc_uci}, where we first normalize covariance matrix of Gaussian copula and covert it by using the relationship between Kendall's tau and Pearson correlation coefficient \cite{MeyerC2013cstm}. The number of cylinders and model year shows discordant pairs due to the 70's oil crisis and horse power and weight shows concordance pair  because more horse power needs bigger engine in general.

\subsection{UCI SPECT Heart Dataset}
\begin{figure}[htp]
    \centering
    \begin{subfigure}[t]{0.4\textwidth}
        \includegraphics[width=\textwidth]{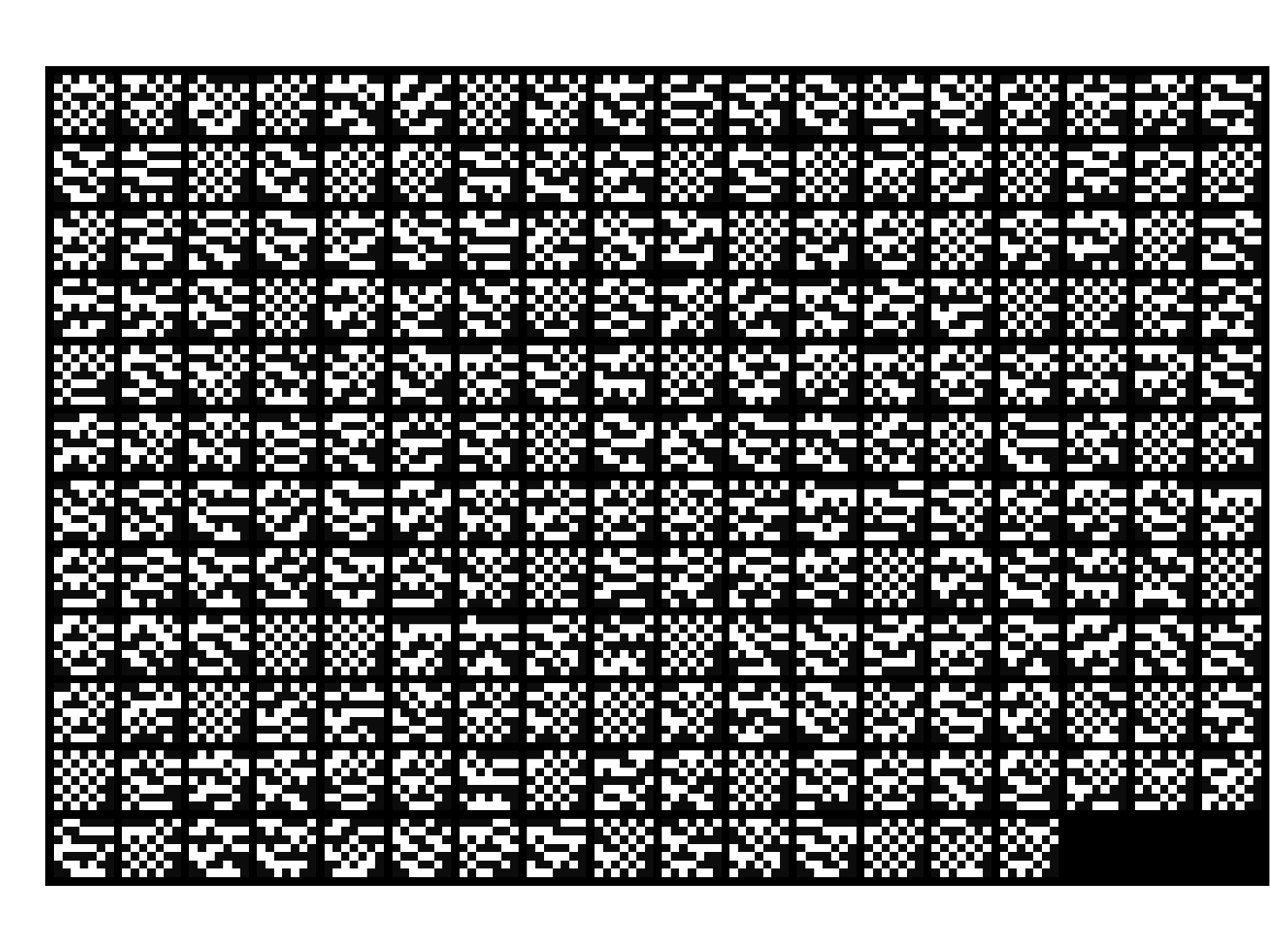}
        \caption{UCI SPECT dataset}
        \label{fig:SPECT_dataset}
    \end{subfigure}\\
    ~
    \begin{subfigure}[t]{0.4\textwidth}
        \includegraphics[width=\textwidth]{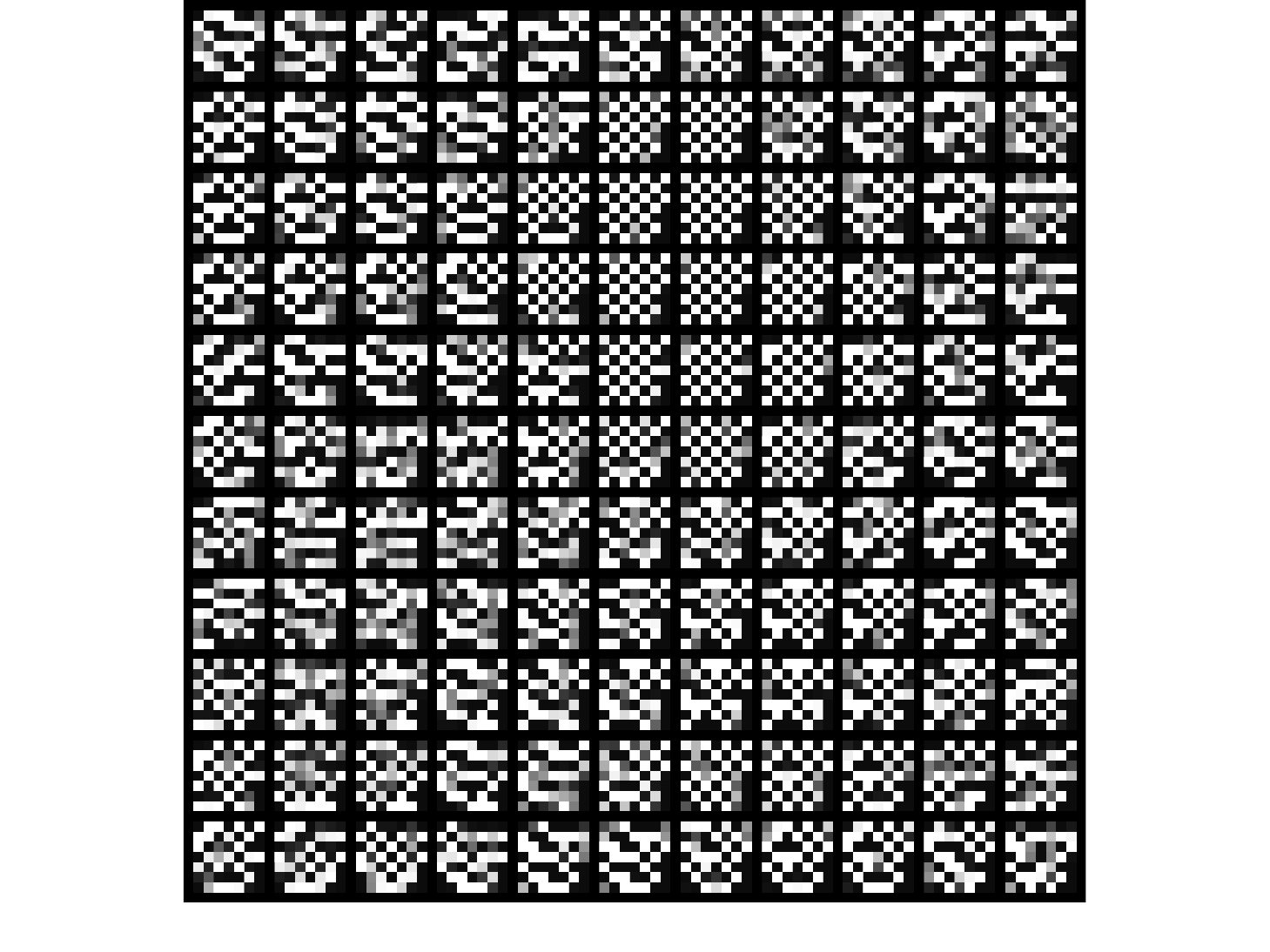}
        \caption{Manifold captured by VAE with 10 $\times$ 10 samples. To show gradual change in manifold, we visualize parameters. }
        \label{fig:SPECT_VAE_manifold}
    \end{subfigure}
    ~ 
    \begin{subfigure}[t]{0.4\textwidth}
        \includegraphics[width=\textwidth]{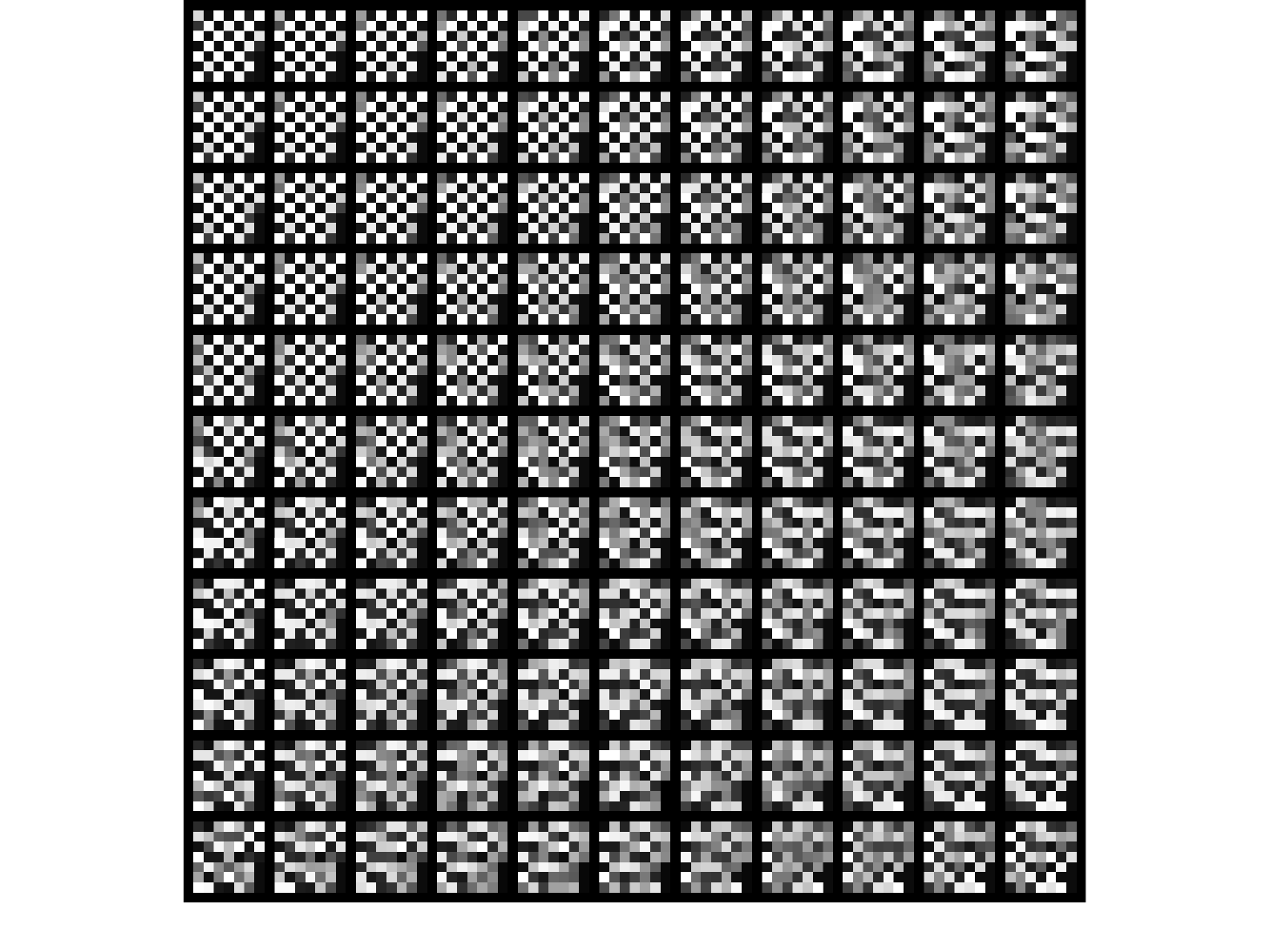}
        \caption{Smooth manifold captured by Gaussian Copula VAE with 10 $\times$ 10 samples}
        \label{fig:SPECT_Copula_VAE_manifold}
    \end{subfigure}
    \caption{VAE and GCVAE on UCI SPECT dataset}\label{fig:uci_auto}
\end{figure}

As shown in Fig. \ref{fig:SPECT_dataset}, UCI SPECT heart dataset describes medical diagnosis on cardiac disease with Single Proton Emission Computed Tomography (SPECT) images. Each instance consists of 22 binary predictors and binary predict that indicates normal and abnormal. We split the total 267 SPECT images into three subset: 213 instances for the training set and 27 instances for validation and test set respectively. After training, both models successfully capture the manifolds in Fig. \ref{fig:SPECT_VAE_manifold} and Fig. \ref{fig:SPECT_Copula_VAE_manifold}. Both models adopt two 100 ReLU hidden units. GCVAE is not regularized by locality constraint. As the same manner as the previous subsection, we fit nonparametric kernel density estimator with the samples  and report the mean and standard deviation of test log-likelihood in Table \ref{tab:ll_copulaVAE} with 10 runs of both models, which shows GCVAE models better than VAE quantitatively.

\section{Conclusions}
\label{sec:conclusions}

In this paper we have presented two extensions of the Gaussian VAE, leading to VAE-ROC and GCVAE.
First, we have developed the VAE-ROC, where the local manifold structure in the unlabelled training dataset
is approximated by a local principal direction at a position determined by the output of a neural network which takes the latent variable $\bz$ as input.
The VAE-ROC was interpreted as an infinite mixture of probabilistic component analyzers, where mixture membership
is smoothly varying, according to latent variables drawn from the standard multivariate Gaussian distribution.
Second, we have developed the GCVAE to handle mixed categorical and continuous data, incorporating Gaussian copula
to model local dependency in the mixed data.
The technique known as 'continuous extension' was used to take care of discrete variables together with continuous variables
in the likelihood of mixed data.
Experiments on various datasets demonstrated the useful behaviour of VAE-ROC and GCVAE, compared to the standard VAE.

\medskip
\noindent
{\bf Acknowledgments:}
This work was supported by Institute for Information \& Communications Technology Promotion (IITP) grant funded 
by the Korea government (MSIP) [B0101-16-0307; Basic Software Research in Human-level Lifelong Machine Learning 
(Machine Learning Center)] and National Research Foundation (NRF) of Korea [NRF-2013R1A2A2A01067464].

\bibliographystyle{unsrt}
\bibliography{/users/seungjin/pub/bib/sjc}

\begin{thebibliography}{10}

\bibitem{KingmaDP2014iclr}
D.~P. Kingma and M.~Welling.
\newblock Auto-encoding variational {Bayes}.
\newblock In {\em Proceedings of the International Conference on Learning
  Representations (ICLR)}, Banff, Canada, 2014.

\bibitem{BengioY2006neco}
Y.~Bengio, M.~Monperrus, and H.~Larochelle.
\newblock Nonlocal estimation of manifold structure.
\newblock {\em Neural Computation}, 18:2509--252, 2006.

\bibitem{NelsenRB2006book}
R.~B. Nelsen.
\newblock {\em An Introduction to Copulas}.
\newblock Springer, 2 edition, 2006.

\bibitem{EverittBS88spl}
B.~S. Everitt.
\newblock A finite mixture model for the clustering of mixed-mode data.
\newblock {\em Statistics \& Probability Letters}, 6(5):305--309, 1988.

\bibitem{HoffPD2007aas}
P.~D. Hoff.
\newblock Extending the rank likelihood for semiparametric copula estimation.
\newblock {\em The Annals of Applied Statistics}, 1(1):265--283, 2007.

\bibitem{DenuitM2005jma}
M.~Denuit and P.~Lambert.
\newblock Constraints on concordance measures in bivariate discrete data.
\newblock {\em Journal of Multivariate Analysis}, 93:40--57, 2005.

\bibitem{MadsenL2011biometrics}
L.~Madsen and Y.~Fang.
\newblock Joint regression analysis for discrete longitudinal data.
\newblock {\em Biometrics}, 67(3), 2011.

\bibitem{LiuH2009jmlr}
H.~Liu, J.~Lafferty, and L.~Wasserman.
\newblock The nonparanormal: Semiparametric estimation of high dimensional
  undirected graphs.
\newblock {\em Journal of Machine Learning Research}, 10:2295--2328, 2009.

\bibitem{MurrayJS2013jasa}
J.~S. Murray, D.~B. Dunson, L.~Carin, and J.~E. Lucas.
\newblock {Bayesian} {Gaussian} copula factor models for mixed data.
\newblock {\em Journal of the American Statistical Association},
  108(502):656--665, 2013.

\bibitem{MarbacM2015arxiv}
M.~Marbac, C.~Biernacki, and V.~Vandewalle.
\newblock Model-based clustering of {Gaussian} copulas for mixed data, 2015.
\newblock {\em Preprint arXiv:1405.1299}.

\bibitem{SklarA59}
A.~Sklar.
\newblock Fonctions de r{\'e}partition {\`a} n dimensions et leurs marges.
\newblock {\em Publ. Inst. Statist. Univ. Paris}, 8:229--231, 1959.

\bibitem{ChorosB2010lns}
B.~Choro{\'s}, R.~Ibragimov, and E.~Permiakova.
\newblock Copula estimation.
\newblock In {\em Copula Theory and Its Applications}, pages 77--91. Springer,
  2010.

\bibitem{MakhzaniA2016iclr}
A.~Makhzani, J.~Shlens, N.~Jaitly, and I.~Goodfellow.
\newblock Adversarial autoencoders.
\newblock In {\em Proceedings of the International Conference on Learning
  Representations (ICLR)}, San Juan, Puerto Rico, 2016.

\bibitem{MagdonIsmailM98nips}
M.~{Magdon-Ismail} and A.~Atiya.
\newblock Neural networks for density estimation.
\newblock In {\em Advances in Neural Information Processing Systems (NIPS)},
  volume~11, 1999.

\bibitem{KingmaDP2015iclr}
D.~P. Kingma and J.~L. Ba.
\newblock {ADAM}: A method for stochastic optimization.
\newblock In {\em Proceedings of the International Conference on Learning
  Representations (ICLR)}, San Diego, CA, USA, 2015.

\bibitem{HintonGE2006neco}
G.~E. Hinton, S.~Osindero, and Y.~W. Teh.
\newblock A fast learning algorithm for deep belief nets.
\newblock {\em Neural Computation}, 18(7):1527--1554, 2006.

\bibitem{GoodfellowIJ2014nips}
I.~J. Goodfellow, J.~{Pouget-Abadie}, M.~Mirza, B.~Xu, D.~{Warde-Farley},
  S.~Ozair, A.~Courville, and Y.~Bengio.
\newblock Generative adversarial nets.
\newblock In {\em Advances in Neural Information Processing Systems (NIPS)},
  volume~27, 2014.

\bibitem{LiY2015icml}
Y.~Li, K.~Swersky, and R.~Zemel.
\newblock Generative moment matching networks.
\newblock In {\em Proceedings of the International Conference on Machine
  Learning (ICML)}, Lille, France, 2015.

\bibitem{DaiZ2016iclr}
Z.~Dai, A.~Damianou, J.~Gonz{\'a}lez, and N.~Lawrence.
\newblock Variational auto-encoded deep {Gaussian} processes.
\newblock In {\em Proceedings of the International Conference on Learning
  Representations (ICLR)}, San Juan, Puerto Rico, 2016.

\bibitem{LeCunY98procieee}
Y.~{LeCun}, L.~Bottou, Y.~Bengio, and P.~Haffner.
\newblock Gradient-based learning applied to document recognition.
\newblock {\em Proceedings of the IEEE}, 86(11):2278--2324, 1998.

\bibitem{BengioY2014icml}
Y.~Bengio, {\'E}.~{Thibodeau-Laufer}, G.~Alain, and J.~Yosinski.
\newblock Deep generative stochastic networks trainable by backprop.
\newblock In {\em Proceedings of the International Conference on Machine
  Learning (ICML)}, Beijing, China, 2014.

\bibitem{AitchisonJ76biometrika}
J.~Aitchison and C.~G.~G. Aitken.
\newblock Multivariate binary discrimination by the kernel method.
\newblock {\em Biometrika}, 63(3):413--420, 1976.

\bibitem{MeyerC2013cstm}
C.~Meyer.
\newblock The bivariate normal copula.
\newblock {\em Communications in Statistics - Theory and Methods},
  42(13):2402--2422, 2013.

\end{thebibliography}

\onecolumn
\begin{appendix}

\section{Gaussian Copula Density with $\bSigma$}
\label{sec:gc}
\begin{proof}
In this section, we derive Gaussian copula with covariance matrix $\bSigma \in \Real^{D \times D}$.
Along with Sklar's theorem, If $ p(\bx)= N(\bzero, \bSigma)$, there exists unique copula function C: $[0,1]^D \rightarrow [0,1]$ satisfying the following condition:
\bee
F(x_1,  \cdots, x_D) &=& C(F_1(x_1), \cdots, F_D(x_D))~,\\
\Phi_{\Sigma}(x_1,  \cdots, x_D) &=& C(\Phi_{\sigma_1}(x_1), \cdots, \Phi_{\sigma_D}(x_D))~,\\
\eee
where $\Phi_{\sigma}$ is the CDF of univariate Gaussian distribution with zero mean and $\sigma^2$ variance and $\Phi_{\Sigma}$ denotes the CDF of multivariate normal distribution of zero mean vector and $\Sigma$ covariance matrix that has $\sigma_d^2$ as $d$th entry of diagonal entries. Inversion of Sklar's theorem, we have:
\be
\label{eq:app_copula}
C_{\Sigma}(u_1, \cdots, u_D) &=& \Phi_{\Sigma}(\Phi^{-1}_{\sigma_1}(u_1), \cdots, \Phi^{-1}_{\sigma_D}(u_D))\\
                                               &=& \Phi_{\Sigma}(\sigma_1\Phi^{-1}(u_1), \cdots, \sigma_D\Phi^{-1}(u_D))\nonumber\\
                                               &=& \Phi_{\Sigma}(q_1, \cdots, q_D)\nonumber\\
                                               &=& \int_{-\infty}^{q_1}\cdots \int_{-\infty}^{q_D} (2\pi)^{-2/D}|\Sigma^{-1/2}|\exp(-\frac{1}{2}\bq^{\top}\bSigma^{-1}\bq) ~dq_1 \cdots dq_D ~,\nonumber
\ee
where $u_d = \Phi_{\sigma_d}(x_d)$ and $\Phi$ is the CDF of univariate standard normal distribution and $q_d = \Phi^{-1}_{\sigma_d}(u_d) = {\sigma_d}\Phi^{-1}(u_d)$ denotes normal score. The second line stands because $\Phi_{\sigma_d}(x_d) = \Phi(x_d/\sigma_d) = u_d$~. By differentiating Eq. \ref{eq:app_copula} with respect $\bu$, we have Gaussian copula density as follows:
\bee
c_{\Sigma}(u_1, \cdots, u_D) &=&  \frac{ \partial^{D} C_{\Sigma}(u_1, \cdots, u_D) }{\partial u_1 \cdots \partial u_D}\\
&=&  \frac{ \partial^{D} C_{\Sigma}(u_1, \cdots, u_D) }{\partial q_1 \cdots \partial q_D}\prod_{d=1}^D \frac{\partial q_d}{\partial u_d}\\
&=& (2\pi)^{-2/D}|\Sigma^{-1/2}|\exp(-\frac{1}{2}\bq^{\top}\bSigma^{-1}\bq)  \prod_{d=1}^D \bigg( \sigma_d \sqrt{2\pi}\exp(\Phi^{-1}(u_d)^2/2)\bigg)\\
&=& \big(\prod_{d=1}^D \sigma_d \big) |\Sigma^{-1/2}|\exp(-\frac{1}{2}\bq^{\top}\bSigma^{-1}\bq)   \exp(\sum_{d=1}^D\frac{1}{2}\Phi^{-1}(u_d)^2)\\
&=& \big( \prod_{d=1}^D \sigma_d \big) |\Sigma^{-1/2}|\exp(\frac{1}{2}\bq^{\top}(  (\mbox{diag}(\bsigma^2))^{-1} - \bSigma^{-1})\bq)~.
\eee
\end{proof}

\section{Sampling from Gaussian copula}
\label{sec:gc_sample}
\begin{figure*}[pth!]
    \centering
    \begin{subfigure}[t]{0.36\textwidth}
        \includegraphics[width=\textwidth]{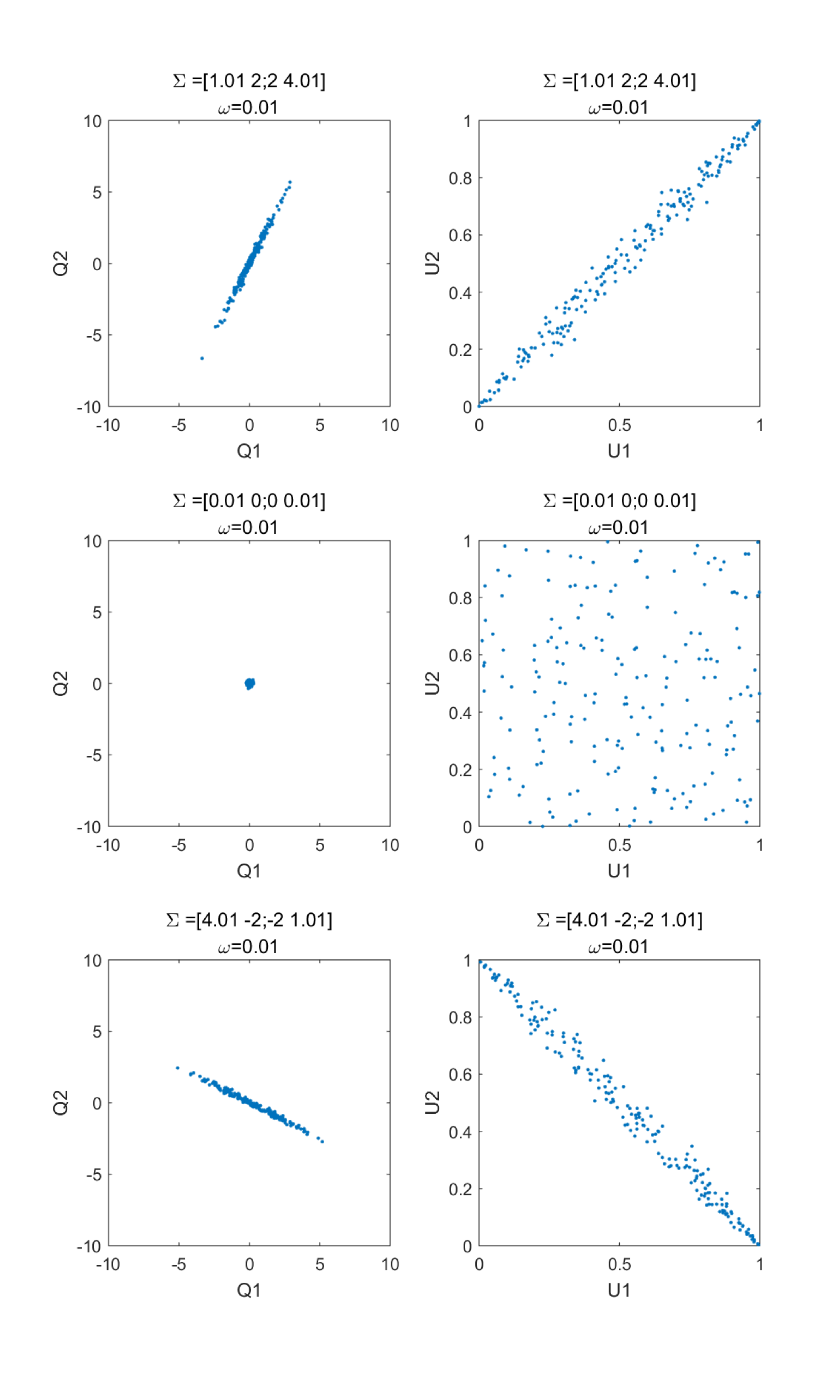}
        \caption{Concordance, No Association, Discordance}
        \label{fig:cs_direction}
    \end{subfigure}
    ~ 
    \begin{subfigure}[t]{0.36\textwidth}
        \includegraphics[width=\textwidth]{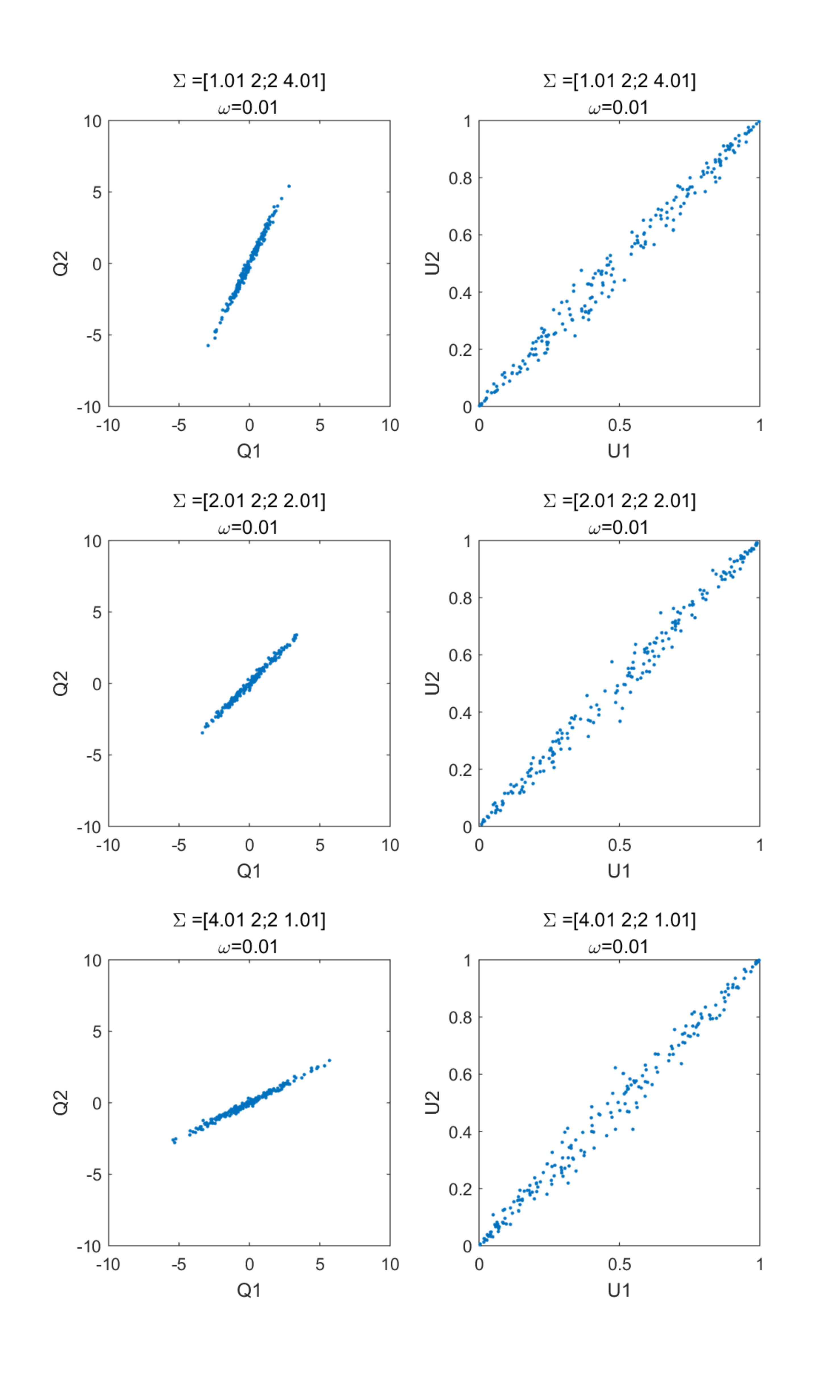}
        \caption{Only concordance matters }
        \label{fig:cs_concord}
    \end{subfigure}
    \\
    \begin{subfigure}[t]{0.36\textwidth}
        \includegraphics[width=\textwidth]{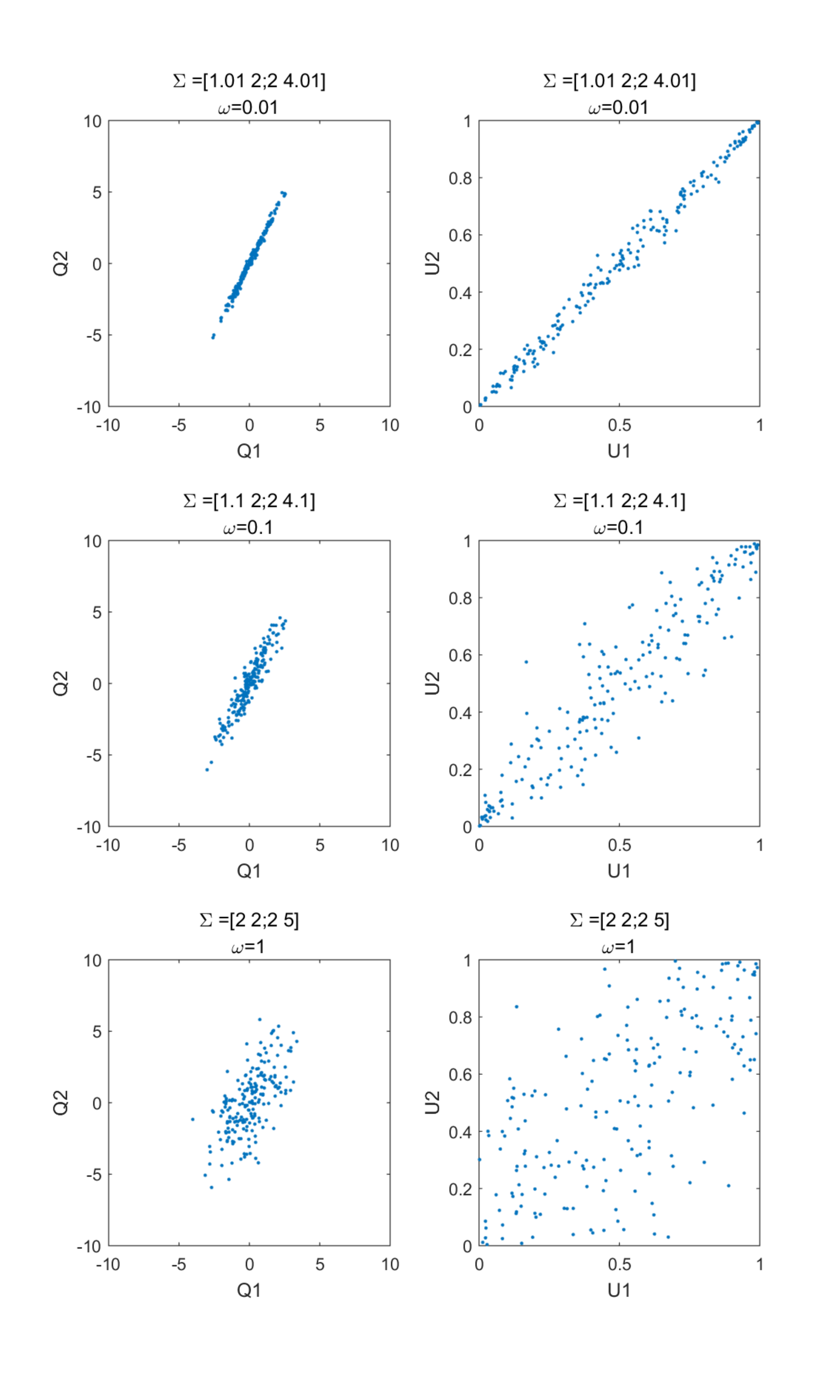}
        \caption{Effect of noise}
        \label{fig:cs_noise}
    \end{subfigure}
    ~
    \begin{subfigure}[t]{0.36\textwidth}
        \includegraphics[width=\textwidth]{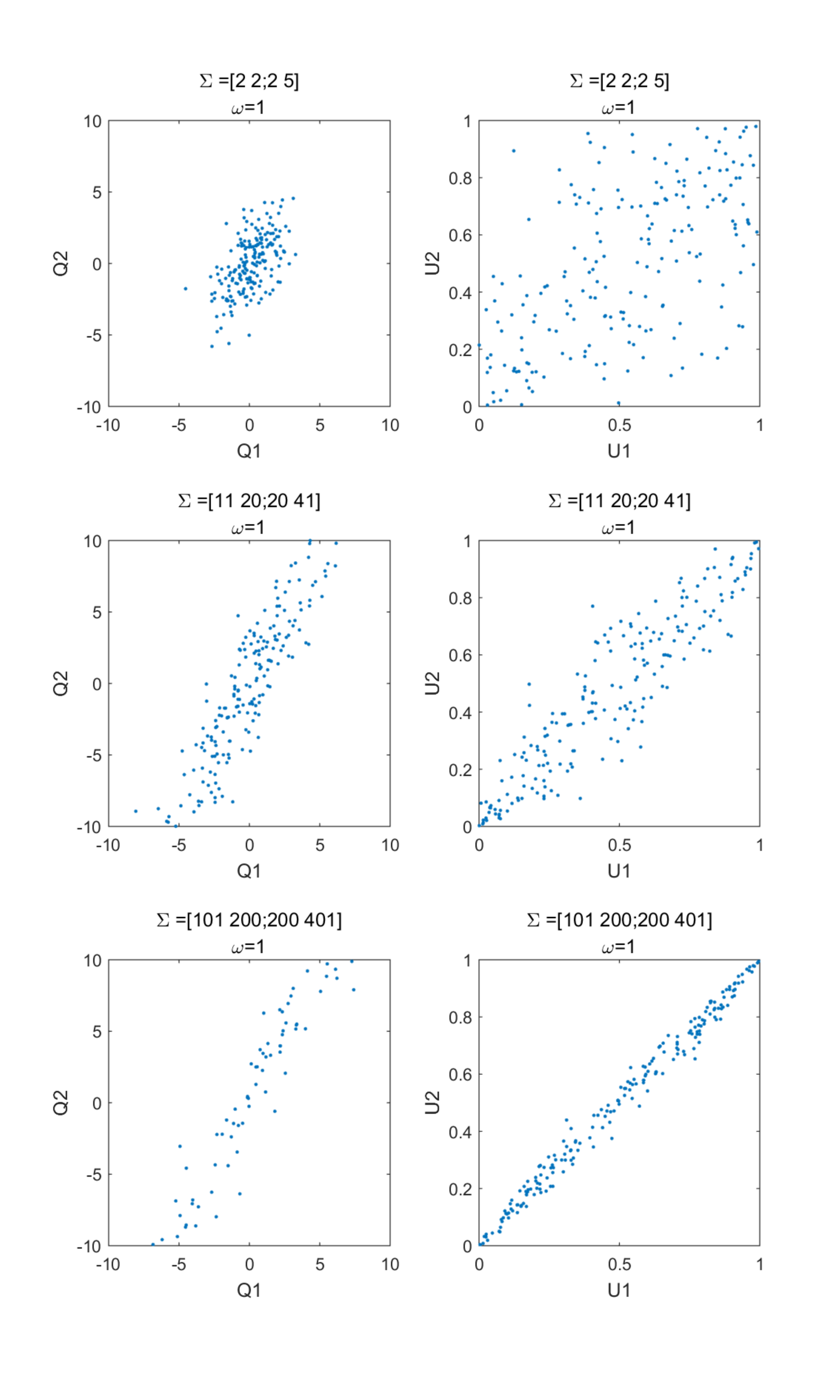}
        \caption{Effect of $||\ba||$  }
        \label{fig:cs_norm}
    \end{subfigure}
    \caption{Samples from Gaussian copula. (left) samples on normal score space  (right)} samples on unit rectangle. \label{fig:sc}
\end{figure*}

In this subsection, we briefly review how to sample from Gaussian copula. Let us assume $\bSigma = \ba\ba^{\top} + \omega \bI$ and we have Gaussian copula having $\bSigma$ as covariance matrix. Then, we have:
\bee
\bq &\sim& \calN(\bzero, \bSigma)~,\\
\bu  &=& [ \Phi\big(\frac{q_1}{\sigma_1}  \big), \cdots,  \Phi\big(\frac{q_D}{\sigma_D}  \big)]^{\top}~.
\eee
By analyzing samples from various cases of parameters $\{ \ba, \omega\}$, we can  more clearly understand how GCVAE models local dependency across input dimensions. Fig. \ref{fig:sc} shows that characteristics of samples with such $\bSigma$ in bivariate Gaussian copula. As shown in Fig. \ref{fig:cs_direction}, If the signs of $a_1$ and $a_2$ are the same, two dimensions have concordant pairs and if they are opposite, two dimensions have discordant pairs. If they are all zeros, there is no dependency between two dimensions. As shown in Fig. \ref{fig:cs_concord}, non-decreasing transformation of any dimension in $\ba$ doesn't change the copula density. Fig. \ref{fig:cs_noise} shows how noise parameter $\omega$ affects copula density function. The large the $\omega$, the lesser dependency two dimensions have. Finally, \ref{fig:cs_norm} shows how norm of principal vector $\ba$ affect the copula density function. The larger the norm is, the more stringent to the noise.

\end{appendix}

\end{document}